\pdfoutput=1

\documentclass[11pt]{article}

\usepackage{ACL2023}

\usepackage{times}
\usepackage{latexsym}
\usepackage[T1]{fontenc}
\usepackage[utf8]{inputenc}
\usepackage{microtype}
\usepackage{inconsolata}

\usepackage{pifont} 
\usepackage{amsmath}
\usepackage{amssymb}
\usepackage{graphicx}
\usepackage{booktabs}
\usepackage{multirow}
\usepackage[capitalize,nameinlink,noabbrev]{cleveref}
\usepackage{bm}
\usepackage{bbm}
\usepackage[shortlabels]{enumitem}
\usepackage{makecell}
\usepackage[super]{nth}
\usepackage{subfig}
\usepackage{balance}
\usepackage{soul}
\usepackage{arydshln}
\usepackage{alltt}
\usepackage{tabu}
\newcommand\rurl[1]{%
  \href{https://#1}{\nolinkurl{#1}}%
}

\newcommand{\citeColored}[2]{\hypersetup{citecolor=#1}\cite{#2}\hypersetup{citecolor=blue}}

\definecolor{pastelblue}{HTML}{A1C9F4}
\definecolor{pastelorange}{HTML}{FFB482}
\definecolor{pastelgreen}{HTML}{8DE5A1}
\definecolor{pastelred}{HTML}{FF9F9B}
\definecolor{pastelpurple}{HTML}{D0BBFF}

\colorlet{pastelbluemuted}{pastelblue!75}
\colorlet{pastelorangemuted}{pastelorange!75}
\colorlet{pastelgreenmuted}{pastelgreen!75}
\colorlet{pastelredmuted}{pastelred!75}
\colorlet{pastelpurplemuted}{pastelpurple!75}

\DeclareRobustCommand{\hlomma}[1]{{\sethlcolor{pastelbluemuted}\hl{#1}}}
\DeclareRobustCommand{\hlommb}[1]{{\sethlcolor{pastelorangemuted}\hl{#1}}}
\DeclareRobustCommand{\hlommc}[1]{{\sethlcolor{pastelgreenmuted}\hl{#1}}}
\DeclareRobustCommand{\hlommd}[1]{{\sethlcolor{pastelpurplemuted}\hl{#1}}}

\DeclareRobustCommand{\hladda}[1]{{\sethlcolor{pastelbluemuted}\hl{#1}}}
\DeclareRobustCommand{\hladdb}[1]{{\sethlcolor{pastelorangemuted}\hl{#1}}}
\DeclareRobustCommand{\hladdc}[1]{{\sethlcolor{pastelgreenmuted}\hl{#1}}}
\DeclareRobustCommand{\hladdd}[1]{{\sethlcolor{pastelpurplemuted}\hl{#1}}}
\DeclareRobustCommand{\hladde}[1]{{\sethlcolor{pastelredmuted}\hl{#1}}}

\makeatletter
\patchcmd{\@@setcref}         {??}{\color{red} undefined Label}{}{}
\patchcmd{\@@setcref}         {??}{\color{red} undefined Label}{}{}
\patchcmd{\@@setcrefrange}    {??}{\color{red} undefined Label}{}{}
\patchcmd{\@@setcrefrange}    {??}{\color{red} undefined Label}{}{}
\patchcmd{\@@setcrefrange}    {??}{\color{red} undefined Label}{}{}
\patchcmd{\@@setcrefrange}    {??}{\color{red} undefined Label}{}{}
\patchcmd{\@@setcrefrange}    {??}{\color{red} undefined Label}{}{}
\patchcmd{\@@setcrefrange}    {??}{\color{red} undefined Label}{}{}
\patchcmd{\@@setnamecref}     {??}{\color{red} undefined Label}{}{}
\patchcmd{\@@setnamecref}     {??}{\color{red} undefined Label}{}{}
\patchcmd{\@@setcpageref}     {??}{\color{red} undefined Label}{}{}
\patchcmd{\@@setcpageref}     {??}{\color{red} undefined Label}{}{}
\patchcmd{\@@setcpagerefrange}{??}{\color{red} undefined Label}{}{}
\patchcmd{\@@setcpagerefrange}{??}{\color{red} undefined Label}{}{}
\patchcmd{\@@setcpagerefrange}{??}{\color{red} undefined Label}{}{}
\patchcmd{\@@setcpagerefrange}{??}{\color{red} undefined Label}{}{}
\patchcmd{\@@setcpagerefrange}{??}{\color{red} undefined Label}{}{}
\patchcmd{\@@cref}            {??}{\color{red} undefined Label}{}{}
\makeatother

\newenvironment{sanseriffontmedium}{\fontsize{7.5}{7.5}\fontfamily{lmss}\selectfont}{\par}
\newenvironment{sanseriffont}{\fontsize{7}{7}\fontfamily{lmss}\selectfont}{\par}
\usepackage{tikz}
\newcommand{\ColoredDot}[1]{\tikz{\fill[fill=#1!85,line width=0pt]  circle(0.8ex);}}
\newcommand*\keystroke[1]{%
  \tikz[baseline=(key.base)]
    \node[%
      draw,
      fill=white,
      rectangle,
      rounded corners=2pt,
      inner sep=1pt,
      line width=0.5pt,
      font=\scriptsize\sffamily
    ](key) {#1\strut}
  ;
}
\title{Guidance in Radiology Report Summarization:\\ An Empirical Evaluation and Error Analysis}

\author{Jan Trienes$^\dag$\quad Paul Youssef$^\ddag$\quad Jörg Schlötterer$^{\dag\P\ddag}$\quad Christin Seifert$^{\dag\ddag}$\\[0.5em]
$^\dag$University of Duisburg-Essen, University Hospital Essen\\
$^{\P}$University of Mannheim\quad $^\ddag$University of Marburg  \\[0.5em]
\texttt{jan.trienes@uni-due.de}\\
\texttt{\{paul.youssef, joerg.schloetterer, christin.seifert\}@uni-marburg.de}}

\begin{document}
\maketitle

\begin{abstract}
Automatically summarizing radiology reports into a concise impression can reduce the manual burden of clinicians and improve the consistency of reporting.
Previous work aimed to enhance content selection and factuality through guided abstractive summarization.
However, two key issues persist.
First, current methods heavily rely on domain-specific resources to extract the guidance signal, limiting their transferability to domains and languages where those resources are unavailable.
Second, while automatic metrics like ROUGE show progress, we lack a good understanding of the errors and failure modes in this task.
To bridge these gaps, we first propose a domain-agnostic guidance signal in form of variable-length extractive summaries.
Our empirical results on two English benchmarks demonstrate that this guidance signal improves upon unguided summarization while being competitive with domain-specific methods.
Additionally, we run an expert evaluation of four systems according to a taxonomy of 11 fine-grained errors.
We find that the most pressing differences between automatic summaries and those of radiologists relate to content selection including omissions (up to 52\%) and additions (up to 57\%).
We hypothesize that latent reporting factors and corpus-level inconsistencies may limit models to reliably learn content selection from the available data, presenting promising directions for future work.
\end{abstract}

\section{Introduction}
The radiology report is an important tool for radiologists to communicate examination results with other clinicians.
Typically, these reports contain three sections: the background section describing the exam and patient context, the findings section providing a detailed description of observations, and the impression section, which concisely summarizes the key findings~\cite{Kahn:2009:Radiology}.
In the clinical process, the impression is of high importance as it informs further treatments.
However, writing the impression can be time-consuming and error-prone, which is why automatic text summarization systems can substantially improve the quality of clinical reporting~\cite{Gershanik:2011:AMIA}.

From a summarization perspective, this task involves both an \emph{extractive} component, where important findings are copied verbatim into the summary, and an \emph{abstractive} component, forming those findings into a concise conclusion taking into account the full report (example in \cref{fig:example}).
Although abstractive methods generate fluent and relevant summaries, they are prone to hallucinations and their output is difficult to control~\cite{Maynez:2020:ACL,Kryscinski:2020:EMNLP,Huang:2020:EMNLP}.
Therefore, current methods for radiology report summarization employ \emph{guided text summarization} to control the summary content through carefully selected guidance signals such as salient ontology terms~\cite{Sotudeh:2020:ACL}, facts~\cite{Zhang:2020:ACL}, and clinical entities~\cite{Hu:2022:ACL}.

\begin{figure}
\setlength\fboxsep{7pt}
\fbox{%
\parbox{0.95\linewidth}{%
\begin{sanseriffontmedium}
\textbf{Background:} Technique: Chest, AP and lateral. Comparison: \_ and \_. History: Weakness and decreased blood sugar with leg swelling and tenderness.\\\vspace{-3pt}%

\textbf{Findings:} The patient is status post coronary artery bypass graft surgery and apparently mitral valve replacement. The heart is mildly enlarged. The mediastinal and hilar contours appear unchanged. \textcolor{Orange}{There is a slight interstitial abnormality, suggestive of a state of very mild congestion, but no new focal opacity.} \textcolor{RoyalBlue}{A left-sided pleural effusion has resolved although mild scarring or atelectasis persists.} Bones are probably demineralized.\\\vspace{-3pt}%

\textbf{Impression:} Findings suggesting mild pulmonary congestion. Resolution of small left-side pleural effusion.%
\vspace{4pt}\hrule\vspace{4pt}
\textbf{BertAbs (unguided)}\\
findings suggesting mild vascular congestion.\\\vspace{-3pt}%

\textbf{GSum Fixed (guidance = $\{ \ColoredDot{Orange} \}$)}\\\vspace{-3pt}%
findings suggest mild vascular congestion.\\

\textbf{GSum Variable (guidance = $\{ \ColoredDot{Orange},  \ColoredDot{RoyalBlue} \}$)}\\
findings suggest mild vascular congestion. resolution of left-sided pleural effusion.
\end{sanseriffontmedium}
}}
\caption{Example radiology report. We guide abstractive summarization with extractive summaries. We propose to adapt the length of the guidance signal to each report rather than using a fixed setting across all reports which helps to accommodate varying target lengths.}
\label{fig:example}
\end{figure}

\begin{figure*}[t]
\centering
\includegraphics[width=\textwidth]{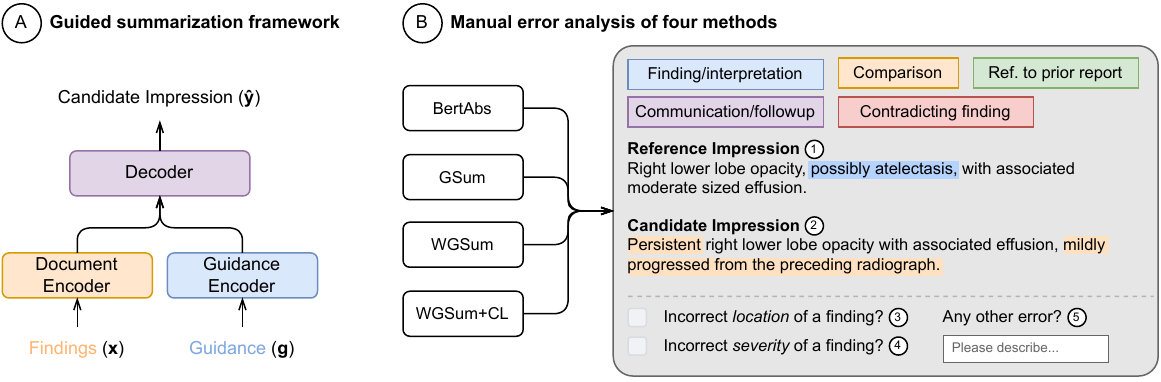}
\caption{Paper overview. \ul{Left:} We evaluate guided methods, where the decoder attends to both the input and a pre-selected guidance signal. \ul{Right:} We task domain experts to identify errors in summaries of unguided and guided methods. Annotation is done on a span-level for omissions from the reference (\ding{192}) and additions to the candidate (\ding{193}), and on an instance-level when both texts report the same finding but with incorrect semantics (\ding{194},\ding{195}). Annotators can flag any other error in free-form (\ding{196}).}
\label{fig:overview}
\end{figure*}

While summary quality has improved steadily, two key issues have received little attention.
First, the success of current methods heavily relies on the availability and quality of the domain-specific guidance extractors (i.e., ontologies, fact extractors, clinical entity taggers).
As these resources are expensive to develop, and as they are only available for a few languages and clinical domains, it is of great interest to investigate to what extent we can use domain-agnostic guidance signals to make guided radiology report summarization methods more easily adoptable.
Second, while we do see improvements in automatic metrics and human assessments of coarse quality criteria such as fluency, correctness and completeness, we lack a good understanding and quantification of the errors and failure modes of current methods. We address the two issues as follows (overview in~\cref{fig:overview}).

\paragraph{Contribution 1: a variable-length extractive guidance signal (\cref{sec:method,sec:results-technical-evaluation}).}
Motivated by the observation that summaries have a large extractive component, we investigate \emph{extractive summaries} as guidance~\cite{Dou:2021:NAACL}.
We identify that the \emph{length of the extractive summaries} is critical for the effectiveness of this guidance signal.
Intuitively, generating longer summaries requires more guidance than shorter ones. 
Therefore, we outline two approaches to adapt the guidance length to each report: (i) a classifier that predicts a suitable length, and (ii) a threshold-based method.
This variable-length guidance signal improves upon unguided summarization, is competitive with recent domain-guided methods, and cheap to adopt as it does not require any domain-specific resources.

\paragraph{Contribution 2: an error analysis (\cref{sec:error-analysis}).}
We conduct an error analysis of unguided, guided and domain-guided methods to identify avenues for improvements of all methods.
We ask domain experts to identify errors in the outputs of four systems and to characterize them along 11 fine-grained categories.
This analysis uncovers three aspects: (1) despite guidance, a significant portion of candidates shows problems in content selection, (2) some content selection decisions are likely only explained by latent factors, (3) there are some dataset-level issues in MIMIC-CXR, including duplicate findings with different impressions, and impression segments without any grounding in the reports.

We make experiment code, full pre-processing pipeline for the datasets, and 1,200 expert assessments of model outputs publicly available.\footnote{\rurl{github.com/jantrienes/inlg2023-radsum}} 

\section{Related Work}

\paragraph{Guided text summarization.}
To address shortcomings in content selection and factuality of neural abstractive summarization methods~\cite{Rush:2015:EMNLP,Nallapati:2016:CoNLL}, guided methods aim to control the content of summaries through carefully selected guidance signals such as keywords~\cite{Li:2018:NAACL}, sentences~\cite{Chen:2018:ACL}, entities~\cite{Fan:2018:ACLws,Narayan:2021:TACL}, templates~\cite{Cao:2018:ACL} and prompts~\cite{He:2022:EMNLP}.

In the radiology domain, \citet{Zhang:2018:LOUHI} proposed to guide generation with the background section of reports using a dual-encoder model.
With a similar architecture, subsequent work explored the use of salient ontology terms~\cite{MacAvaney:2019:SIGIR,Sotudeh:2020:ACL}.
Besides text-based guidance, \citet{Hu:2021:ACL,Hu:2022:ACL} propose a graph-guided decoder which attends both to the report text and to a word-graph of clinical entities.
In contrast, we explore extractive summaries as guidance signal~\cite{Liu:2019:EMNLP,Dou:2021:NAACL}.
Crucially, this guidance signal can be extracted without any domain-specific resources such as ontologies and clinical entity taggers.
To this end, our method is similar to the approach by~\citet{Zhu:2023:arXiv}, which guides summarization with reference summaries from the training set, selected at random or by an oracle. Similar to our approach, this guidance signal can vary in length.

\paragraph{Alternatives to guided summarization.}
Several parallel research lines aim to enhance radiology report summarization with a different methodological focus.
First, several studies optimize factual consistency through reinforcement learning~\cite{Zhang:2020:ACL,Delbrouck:2022:ML4H} or post-hoc reranking~\cite{Xie:2023:arXiv}.
Second,~\citet{Karn:2022:ACL} devise an extract-then-abstract pipeline with multi-agent reinforcement learning.
Last, recent work explores domain-adaptation techniques for pre-trained language models to better accommodate radiology reports~\cite{Cai:2021:TransMultimedia,VanVeen:2023:ACL-ws}.
Our work is orthogonal to these efforts and future work could investigate how to combine them with guided summarization.

\paragraph{Evaluation of radiology report summarization.}
Evaluating text summarization systems is a long standing issue.
Since automatic metrics have a limited correlation with human judgment~\cite{Fabbri:2021:TACL}, manual evaluation is still regarded as the gold standard.
For the task of radiology report summarization, most manual evaluations focus on coarse criteria such as accuracy, completeness, conciseness, and readability~\cite{Zhang:2018:LOUHI,Hu:2022:ACL,Cai:2021:TransMultimedia}.
Yet, these evaluations only provide limited insights into directions for improvement.
To support the interpretation of automatic and manual evaluations, and to understand the pitfalls of current methods, we conduct an error analysis~\cite{Van-Miltenburg:2021:INLG}.
In this line of work,~\citet{Yu:2022:medRxiv} evaluated the ability of automatic metrics to capture six fine-grained errors of radiograph-to-impression models.
We extend this taxonomy in our error analysis.

\section{Method}
\label{sec:method}
We formulate the task of summarizing radiology reports as follows.
Given the findings section of a report, represented as a sequence of tokens $\bm{x} = (x_1, x_2,\ldots, x_N)$, the goal is to generate an impression $\bm{y} = (y_1,y_2,\ldots,y_M)$ that accurately summarizes the most significant findings.
The guided summarization framework extends this problem setting with an additional input signal $\bm{g} = (g_1, g_2,\ldots, g_L)$ which aims to improve the quality of generated summaries by indicating salient information in $\bm{x}$.

\subsection{Model and Extractive Guidance}
\label{sec:method-extractive-guidance}
As a concrete implementation of the guided summarization framework, we adopt GSum~\cite{Dou:2021:NAACL}.
This sequence-to-sequence model extends a transformer-based architecture for abstractive text summarization~\cite{Liu:2019:EMNLP} with an additional encoder for the guidance signal $\bm{g}$.
To create a guidance-aware representation of the input, the decoder first attends to the encoded representation of $\bm{g}$, and afterwards to the whole input document $\bm{x}$ using cross-attention~\cite{Vaswani:2017:NIPS}.
The authors demonstrate that GSum is effective at controlling the content of summaries, leading to good results on several non-medical datasets.

\paragraph{Extractive guidance.}
While $\bm{g}$ can take any form, \citet{Dou:2021:NAACL} found the output of an extractive summarization to be highly effective.
Intuitively, this guidance signal informs the model about which input sentences should be highlighted in a summary.
An important implementation detail of GSum is the mechanism to obtain the extractive sentences.
\citet{Dou:2021:NAACL} distinguish between the \textbf{oracle} setting and the \textbf{automatic} setting.
In the oracle setting, the guidance sentences are greedily picked from $\bm{x}$ such that they maximize \textsc{rouge} with respect to $\bm{y}$~\cite{Nallapati:2017:AAAI}.
In the automatic setting, this oracle is approximated by an extractive summarization method (BertExt,~\citealp{Liu:2019:EMNLP}).
The training labels for BertExt are derived using the same greedy matching, thus BertExt can be considered an approximation of the oracle guidance.
Selecting the guidance signal from BertExt follows a top-$k$ strategy: scoring all sentences for relevance and selecting the highest scoring sentences until a predefined length threshold is reached~\cite{Nallapati:2017:AAAI,Liu:2019:EMNLP}.
Following \citet{Dou:2021:NAACL}, the oracle signal is used during training of GSum.
During inference, we state explicitly whether we use the oracle or automatic variant.

\subsection{Variable-length Extractive Guidance}
\label{sec:method-variable-length-guidance}
We empirically find that extracting a fixed-length summary with the top-$k$ approach has a negative impact on the effectiveness of GSum (\cref{sec:results-gsum-fixed}).
To address this problem, we propose two methods to select a variable-length extractive guidance signal from BertExt.
Formally, for a given document $\bm{x}$ and its sequence of sentences $(s_1,\ldots,s_N)$, with $s_{i}$ being the $i$-th sentence in $\bm{x}$, these methods have to select $L<N$ sentences as guidance $\bm{g}$.

\paragraph{Method 1: predicting oracle length.}
As described in \cref{sec:method-extractive-guidance}, BertExt is trained to assign a label $y \in \{0, 1\}$ to each sentence $s_{i}$. The predicted probability $p(y = 1|s_{i})$ indicates if $s_{i}$ should be included in the summary.
The ground-truth labels are derived from an extractive oracle $f_{\text{oracle}}(\bm{x}, \bm{y})$ which greedily selects a subset of sentences of length $[0,3]$ that maximizes \textsc{rouge} against the gold summary $\bm{y}$~\cite{Liu:2019:EMNLP}.
Instead of taking a fixed number of sentences with highest probability (top-$k$), we train a sequence-classification model to predict the length of the extractive oracle $f_{\text{approx}}(\bm{x}) = L \in [0,3]$, and select the top-$L$ sentences as guidance signal.

\paragraph{Method 2: threshold calibration.}
Instead of considering the full ranked list of sentences, this method constrains selection with a threshold-based approach inspired by \citet{Jia:2021:AAAI}.
Recall that $p(y = 1|s_{i})$ denotes the probability that BertExt assigns to the positive class.
We then select the set of sentences that exceed a probability threshold $T$ as guidance signal:
\begin{equation*}
  \bm{g} = \{s_{i} \in \bm{x} | p(y=1|s_{i}) \ge T \}.
\end{equation*}
We optimize $T \in [0,1]$ on a validation set to maximize \textsc{rouge}-1.

\section{Technical Evaluation}
\label{sec:results-technical-evaluation}

\noindent\fbox{%
\parbox{0.98\linewidth}{%
\fontsize{9.5}{11.5}\selectfont
\textbf{RQ1.} To what extent are extractive summaries an effective guidance signal for radiology report summarization?\\[0.5em]
\textbf{RQ2.} How does adapting the extractive guidance length to each report impact the overall quality of summaries?
}}

\subsection{Experimental Setup}
\label{sec:experimental-setup}

\paragraph{Datasets.}
\label{sec:datasets}
We use two public datasets of English chest x-ray reports: \textbf{MIMIC-CXR}~\cite{Johnson:2019:SciData} and \textbf{OpenI}~\cite{Demner-Fushman:2015:JAMIA}.
Consistent with prior work~\cite{Zhang:2018:LOUHI,Sotudeh:2020:ACL,Hu:2022:ACL}, we retain reports with exactly one findings and one impression section, where both have an acceptable length ($\ge 10$ tokens in findings, $\ge 2$ tokens in impression), and we discard the background section.\footnote{%
  To compare the relative utility of guidance signals, including the background section is not necessary. For completeness, we report results with background section in \cref{sec:error-analysis-discussion}.
}
Following \citet{Hu:2022:ACL}, we use the official training, validation and test splits of MIMIC-CXR and a random split with a 70/10/20 ratio for OpenI.
We use \textsc{spacy} for tokenization and \textsc{nltk} for sentence segmentation.\footnote{\rurl{spacy.io} and \rurl{nltk.org}}
\cref{tab:datasets} reports the dataset statistics.

\begin{table}[t]
\small
\centering
\resizebox{\columnwidth}{!}{
\begin{tabular}{lcc}
\toprule
\textbf{Aspect} & \textbf{MIMIC-CXR} & \textbf{OpenI} \\
\midrule
Reports & 122,500$\,$/$\,$963$\,$/$\,$1,598 & 2,342$\,$/$\,$334$\,$/$\,$670 \\
Avg. $|\bm{x}|_{t}$ & 56 {\color{gray} $\pm$ 25.2} & 37 {\color{gray} $\pm$ 16.4} \\
Avg. $|\bm{x}|_{s}$ & 5.5 {\color{gray} $\pm$ 1.9} & 4.6 {\color{gray} $\pm$ 1.6} \\
Avg. $|\bm{y}|_{t}$ & 15 {\color{gray} $\pm$ 13.5} & 8 {\color{gray} $\pm$ 8.1} \\
Avg. $|\bm{y}|_{s}$ & 1.6 {\color{gray} $\pm$ 0.9} & 1.4 {\color{gray} $\pm$ 0.8} \\
Novelty & 73.4\% & 86.8\% \\
Compression & 73.8\% & 76.1\% \\
\bottomrule
\end{tabular}
}
\caption{%
Statistics of the benchmark datasets, including the number of reports (train/valid/test), length and standard deviation in tokens/sentences ($|\cdot|_{t}$ and $|\cdot|_{s}$), novelty as average percentage of bigrams in impression $\bm{y}$, but not in findings $\bm{x}$, and average compression ($\frac{|\bm{y}|_{t}}{|\bm{x}|_{t}}$).
}
\label{tab:datasets}
\end{table}

\paragraph{Baselines.}
We compare with three categories of baselines: (1) unguided methods, (2) vanilla GSum with fixed-length extractive guidance~\cite{Dou:2021:NAACL}, and (3) domain-specific guided methods.
Regarding unguided methods, we use \textbf{OracleExt}~\cite{Nallapati:2017:AAAI} which greedily selects sentences from the findings that maximize \textsc{rouge} with respect to the impression.
Furthermore, we use \textbf{BertExt} and \textbf{BertAbs}~\cite{Liu:2019:EMNLP} which are extractive/abstractive transformer-based models initialized with pre-trained \textsc{bert}~\cite{Devlin:2019:NAACL}.
Regarding domain-specific methods, we compare with \textbf{WGSum}~\cite{Hu:2021:ACL} which employs a graph-guided decoder to attend to a graph of clinical entities extracted with Stanza~\cite{Zhang:2021:JAMIA}, and with
\textbf{WGSum+CL}~\cite{Hu:2022:ACL} which refines this guidance signal through contrastive learning.

\paragraph{Automatic evaluation metrics.}
We evaluate the quality of generated impressions with \textsc{rouge} $F_1$~\cite{Lin:2004:WS} to measure unigram and bigram overlap as a proxy for relevance (\mbox{R-1}, \mbox{R-2}) and the longest common subsequence as a proxy for fluency (\mbox{R-L}).
In addition, we report BERTScore as a measure of soft-alignment~\cite{Zhang:2020:ICLR}.
As factual correctness is critical, we also calculate a factuality $F_1$~\cite[Fact.]{Zhang:2020:ACL,Hu:2022:ACL}.
This metric is based on a rule-based fact-extraction method, CheXpert~\cite{Irvin:2019:AAAI}, which labels the status (present, absent, uncertain) of 14 radiological observations.
By applying this procedure to both the reference and candidate summary, we can calculate a precision/recall of facts.

\paragraph{Implementation and hyperparameters.}
For all summarization models, we use the hyperparameters and code of the original papers.
Below, we focus on deviations from those settings and report all hyperparameters in \cref{sec:appendix-replication-models}.

For BertExt, BertAbs and GSum, we make three adaptations: (i) the summary length of BertExt is set to the average number of sentences selected by OracleExt, rounded to the nearest integer,\footnote{On MIMIC-CXR and OpenI $|\text{OracleExt}(\bm{x}, \bm{y})| \approx 1$.} (ii) we reduce the training steps to 20,000 to account for the smaller datasets, and (iii) to address an exploding gradient problem, we reduce the initial learning rate by a factor of 10.
For final testing, we take the checkpoint with lowest validation loss on MIMIC-CXR. On OpenI, we found the loss to be unstable, so opted to select models by validation R-1.

Regarding the guidance-length prediction models (Method 1 in \cref{sec:method-variable-length-guidance}), we experiment with two classifiers.
First, a multinomial logistic regression classifier with unigram bag-of-words features (\textsc{lr-approx}).
Second, as this model may be too simplistic to accurately predict the guidance length, we implement a transformer-based classifier (\textsc{bert-approx}) on top of DistilBERT~\cite{Sanh:2019:DistilBERT}.

\subsection{Fixed-length Guidance (RQ1)}
\label{sec:results-gsum-fixed}
We first aim to understand if extractive summaries can be a useful guidance signal for radiology report summarization.
To this end, we compare BertAbs (i.e., unguided) with GSum in its default configuration (Part 1 in \cref{tab:results}).

\begin{table*}[t]
\small
\centering
\begin{tabu}{lcccccccccc}
\toprule
{} & \multicolumn{5}{c}{\textbf{MIMIC-CXR}} & \multicolumn{5}{c}{\textbf{OpenI}} \\
\cmidrule(lr){2-6}
\cmidrule(lr){7-11}
\textbf{Method} & R-1 & R-2 & R-L & BS & Fact. & R-1 & R-2 & R-L & BS & Fact. \\
\midrule
\multicolumn{11}{l}{\emph{Part 1: Baselines and reproduction of GSum}} \\
OracleExt & 44.0 & 25.4 & 40.6 & 50.1 & 55.1 & 30.5 & 11.9 & 29.2 & 33.7 & 53.5 \\
BertExt~\cite{Liu:2019:EMNLP} & 32.7 & 18.1 & 30.0 & 41.9 & 44.5 & 23.6 & 7.4 & 22.6 & 32.2 & \textbf{42.8} \\
BertAbs~\cite{Liu:2019:EMNLP} & 48.4 & 34.1 & 46.6 & 58.8 & 47.3 & 62.0 & 52.7 & 61.7 & 69.2 & 39.3 \\
GSum~\cite{Dou:2021:NAACL} & 46.3 & 32.7 & 44.7 & 57.4 & 46.6 & 60.1 & 49.6 & 59.8 & 67.0 & 40.0 \\

\addlinespace\multicolumn{11}{l}{\emph{Part 2: GSum adapted with a variable-length guidance signal (ours)}} \\
GSum w/ LR-Approx & 48.9 & 34.2 & 47.0 & 59.1 & 48.2 & 62.0 & 51.2 & 61.6 & 67.9 & 41.7 \\
GSum w/ BERT-Approx & 49.4 & 34.5 & 47.4 & 59.5 & 50.6 & 62.5 & 51.6 & 62.2 & 68.4 & 39.6 \\
GSum w/ Thresholding & \textbf{49.9} & 34.3 & \textbf{47.8} & \textbf{59.8} & 49.0 & 62.2 & 50.8 & 61.8 & 68.6 & 40.4 \\

\addlinespace\multicolumn{11}{l}{\emph{Part 3: Comparison with domain-specific methods}} \\
WGSum~\cite{Hu:2021:ACL} & 48.4 & 32.8 & 46.5 & 58.6 & 49.8 & 61.1 & 50.0 & 60.8 & 67.9 & 38.4 \\
WGSum+CL~\cite{Hu:2022:ACL} & 49.5 & \textbf{35.3} & 47.8 & 59.5 & \textbf{51.1} & \textbf{64.7} & \textbf{57.1} & \textbf{64.5} & \textbf{70.0} & 37.2 \\
\rowfont{\color{gray}} WGSum~\citeColored{gray}{Hu:2021:ACL}$^\dagger$ & 48.3 & 33.3 & 46.6 & --- & --- & 61.6 & 50.9 & 61.7 & --- & --- \\
\rowfont{\color{gray}} WGSum+CL~\citeColored{gray}{Hu:2022:ACL}$^\dagger$ & 49.1 & 33.7 & 47.1 & --- & --- & 64.9 & 55.5 & 64.4 & --- & --- \\
\bottomrule
\end{tabu}
\caption{Technical evaluation of unguided, guided and domain-guided methods on two datasets. Metrics are ROUGE-1/2/L, BERTScore (BS) and CheXpert factuality F1 (Fact). All results were obtained by re-implementing
the models with the official code of respective papers, results directly cited are indicated with $\dagger$.}
\label{tab:results}
\end{table*}

We find that \textbf{GSum with fixed-length extractive guidance~\cite{Dou:2021:NAACL} does not generalize to the radiology domain.}
Compared with BertAbs, effectiveness decreases by 4.5\% and 3.2\% in R-1 for MIMIC-CXR and OpenI, respectively.
This is surprising as GSum is highly effective on multiple non-medical summarization benchmarks under the same experimental conditions~\cite{Dou:2021:NAACL}.
Our hypothesis is that highly varying summary lengths make the standard fixed-length guidance in GSum ineffective on this data.\footnote{\cref{sec:appendix-target-summary-length} gives the length distribution of targets.}
We empirically verify this hypothesis in the following experiments.

\paragraph{Comparing oracle and automatic guidance.}
To get an upper-bound estimate for extractive guidance signals, we analyze GSum in an unrealistic oracle setting.
Recall from \cref{sec:method-extractive-guidance} that during training of GSum, the guidance signal is extracted by OracleExt, whereas during inference guidance is extracted by BertExt with a summary length fixed to $k=1$ across all reports.
If we instead also use OracleExt as guidance extractor during inference, we see a substantial increase in all metrics (R-1 46.3\textrightarrow58.8 on MIMIC, and R-1 60.1\textrightarrow68.8 on OpenI, all metrics in Appendix \cref{tab:appendix-gsum-oracle}).
\textbf{This oracle experiment demonstrates (i) that GSum learned to rely on guidance, and (ii) that extractive summaries can be a highly effective guidance signal if selected in the right way.}

Given that GSum is effective when we use the oracle guidance (OracleExt), it is important to understand how this guidance signal differs from the automatically extracted guidance (BertExt).
We find that a characterizing difference between the two guidance signals is the length of the resulting summaries.
OracleExt produces summaries with 0/1/2/3 sentences for 2/52/32/14\% of the MIMIC-CXR reports, and for 15/67/14/3\% of the OpenI reports.
This implies that a guidance signal with a length of $k=1$ is too short for 46\% of the MIMIC-CXR reports, whereas on OpenI it is too short for 17\% and too long for 15\%.

\subsection{Variable-length Guidance (RQ2)}
\label{sec:results-variable-length-guidance}
We next evaluate the utility of our proposed variable-length extractive guidance signal (Part 2 of \cref{tab:results}). We make several observations.

First, we find that variable-length extractive guidance substantially improves the effectiveness of GSum.
On MIMIC-CXR, our adaptation is also better than unguided summarization (BertAbs).
In particular, we observe a large increase in factuality, which is critical in the clinical domain.
While we see a similar improvement of GSum on OpenI, this guided summarization model does not improve over BertAbs.
One potential reason is that OpenI is more abstractive than MIMIC-CXR, as indicated by the high degree of novelty (\cref{tab:datasets}) and the relatively low scores of the extractive methods (BertExt, OracleExt in \cref{tab:results}).
This corroborates the findings by \citet{Dou:2021:NAACL}, where GSum was less effective on more abstractive datasets.
For future work, it would be interesting to study the interplay between the degree of abstraction, and the utility of extractive guidance signals.

Second, regarding the different strategies to obtain variable-length extractive summaries, we cannot conclude that one is superior over another.
The classifier-based approaches (LR-Approx, BERT-Approx), and the thresholding-based approach (Thresholding) lead to similar results when the extracted guidance is used downstream in GSum.
For each guidance extraction strategy, we calculate the ROUGE scores of the guidance signal with respect to the gold summaries.
From \cref{tab:results-bertex}, we see that all strategies have the desired effect of increasing content recall, with a smaller sacrifice in precision.

\begin{table}[t]
\small
\centering
\resizebox{\columnwidth}{!}{
\begin{tabular}{lr @{\hspace{1.1\tabcolsep}} rr @{\hspace{1.1\tabcolsep}} r}
\toprule
\multirowcell{2}[-0.5ex][l]{\textbf{BertExt}\\\textbf{length} ($k = \cdot$)} & \multicolumn{2}{c}{\textbf{MIMIC-CXR}} & \multicolumn{2}{c}{\textbf{OpenI}} \\
\cmidrule(lr){2-3}
\cmidrule(lr){4-5}
& R-1 (Prec./Rec.) & $|\hat{\bm{y}}|$ & R-1 (Prec./Rec.) & $|\hat{\bm{y}}|$ \\
\midrule
Fixed ($k = 1$) & 32.7 (\textbf{38.5}/34.2) & 1.0 & \textbf{23.6} (\textbf{24.6}/26.9) & 1.0 \\
\textsc{lr-approx} & 34.5 (35.7/40.0) & 1.4 & 23.5 (23.9/27.2) & 1.1 \\
\textsc{bert-approx} & 35.2 (34.6/42.0) & 1.5 & 23.5 (23.7/27.5) & 1.1 \\
Thresholding & \textbf{36.1} (34.1/\textbf{46.3}) & 1.7 & 23.2 (22.9/\textbf{29.0}) & 1.2 \\
\midrule
$k = |\text{OracleExt}|$ & 36.9 (35.3/44.2) & 1.6 & 24.3 (23.2/29.2) & 1.2 \\
\bottomrule
\end{tabular}
}
\caption{
Comparing strategies for extracting variable-length summaries with BertExt by measuring \textsc{rouge} against the gold summary.
Average summary length $|\hat{\bm{y}}|$ given in sentences.
All methods are tested as guidance signal for GSum in \cref{tab:results}.
}
\label{tab:results-bertex}
\end{table}

\begin{figure}
\centering
\includegraphics[width=.9\columnwidth]{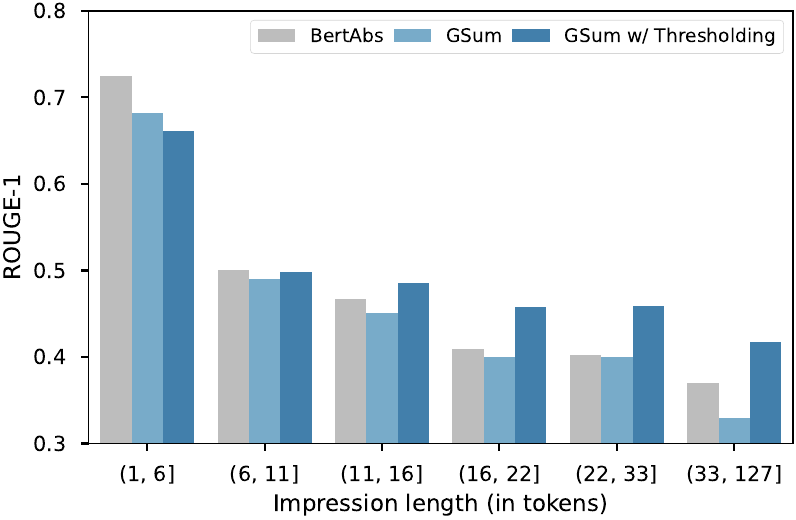}
\caption{Evaluating summaries by target length on MIMIC-CXR (equal number of samples per bucket).}
\label{fig:results-rouge-by-target}
\end{figure}

Third, to better understand how guidance influences the quality of summaries, we plot the \mbox{R-1} scores across different target summary lengths (\cref{fig:results-rouge-by-target}).
We find that variable-length guidance improves the quality of longer summaries, while for shorter targets, extractive guidance is not beneficial.
By manual inspection, we find that short targets are standard phrasings of negative results (e.g., \emph{``No evidence of acute findings''}), whereas longer targets have a higher extractive component by reiterating particular findings.
In practice, it could be interesting to combine unguided and guided methods by letting the radiologist decide whether a long or short summary should be generated.

\paragraph{Comparison with domain-specific guided summarization (WGSum, WGSum+CL).}
Lastly, compared with the domain-specific guided methods (Part 3 of \cref{tab:results}), we find on MIMIC-CXR that GSum with variable-length extractive guidance is just as effective as WGSum and WGSum+CL which use a graph of clinical entities.
On OpenI, our approach improves over WGSum, but is slightly worse than WGSum+CL.

\paragraph{Summary of RQ1/RQ2.}
Overall our results show that extractive summaries are a promising guidance signal for clinical reports without requiring any domain-specific resources.
We envision that this makes it easier to adopt guided summarization in other clinical domains and languages, for which domain-specific resources like ontologies and clinical NER models are not widely available.

\section{Error Analysis}
\label{sec:error-analysis}
\noindent\fbox{%
\parbox{0.98\linewidth}{%
\fontsize{9.5}{11.5}\selectfont
\textbf{RQ3.} What are the errors and failure modes of unguided and guided methods for radiology report summarization?
}}

\subsection{Evaluation Setup}
Inspired by the Multidimensional Quality Metrics framework for evaluation of machine translation systems (\citealp{Lommel:2014:MQM}), we conduct a span-based error annotation.
We task annotators to highlight erroneous text spans and to classify them according to an error taxonomy.
As a starting point, we use the taxonomy proposed by \citet{Yu:2022:medRxiv}.
Based on two pilot runs, we extended this taxonomy from initially 6 to 11 fine-grained error categories (see \cref{tab:error-analysis-results}) and developed a definition and examples for each. 
Following \citet{Yu:2022:medRxiv}, we opt for a reference-based evaluation.
We want to understand how the system generated summary differs from the clinician summary both in content and correctness of the presented facts.
Therefore, our errors can be grouped into additions (spans in the candidate), omissions (spans in the reference), and binary choices for the correctness of presented facts.
Further, we ask annotators to flag any additional errors they encounter as a free-form answer.
We provide full annotation guidelines in~\cref{sec:appendix-annotation-guidelines}.

\paragraph{Materials.}
We randomly select 100 reports from the official test set of MIMIC-CXR which is stratified to cover both frequent and less frequent inputs/clinical observations~\cite{Johnson:2019:arXiv}.
For each input, we generate four candidate summaries using BertAbs (representative of unguided systems), GSum w/ Thresholding (representative of systems with domain-agnostic guidance), and WGSum/WGSum+CL (representative of systems with domain-specific guidance).
We present the reference summary and all candidates (in random order) at once to annotators to reduce effort and ensure consistent annotation of similar summaries.
Each set of summaries is completed by three annotators resulting in 1,200 error annotations (100~reports \texttimes\ 4~candidates \texttimes\ 3~annotators). We form a ``gold standard'' from the triple annotation by majority voting (example aggregation in \cref{sec:appendix-span-based-aggregation}).

\paragraph{Annotators.}
To account for the domain knowledge necessary for the annotation task, we hired 6 senior medical students in their fifth year of training. All annotators are fluent in English. We compensated annotators with 10.5€ per hour (standard rate for student assistants in Germany). The annotation took 23.1 hours (avg. 4.6 min/sample), plus additional time for pilot rounds and discussions.

\begin{figure*}[t]
\small
\resizebox{\columnwidth}{!}{
\begin{tabular}{
    l @{\hspace{1\tabcolsep}} l @{\hspace{1\tabcolsep}}
    r @{\hspace{0.3\tabcolsep}} >{\color{gray}}r @{\hspace{1\tabcolsep}}
    r @{\hspace{0.3\tabcolsep}} >{\color{gray}}r @{\hspace{1\tabcolsep}}
    r @{\hspace{0.3\tabcolsep}} >{\color{gray}}r @{\hspace{1\tabcolsep}}
    r @{\hspace{0.3\tabcolsep}} >{\color{gray}}r @{\hspace{1\tabcolsep}}
}
\toprule
\textbf{\#} & \textbf{Error Category} & \textbf{M1} & (\%) & \textbf{M2} & (\%) & \textbf{M3} & (\%) & \textbf{M4} & (\%) \\
\midrule
0 & No error & 20 & (20) & 18 & (18) & 14 & (14) & \textbf{22} & (22) \\

\addlinespace\multicolumn{10}{l}{\emph{Omissions from reference}} \\
1a & \hlomma{Finding/interpretation} & 70 & (52) & \textbf{58} & (43) & 62 & (48) & 64 & (47) \\
1b & \hlommb{Comparison} & 23 & (19) & \textbf{16} & (15) & 19 & (16) & 23 & (19) \\
1c & \hlommc{Ref. to prior report} & \textbf{1} & (1) & 3 & (3) & 2 & (2) & 2 & (2) \\
1d & \hlommd{Communication/followup} & 20 & (19) & \textbf{18} & (16) & 19 & (17) & 19 & (17) \\
\midrule
\multicolumn{2}{l}{Total} & 114 & (66) & \textbf{95} & (58) & 102 & (63) & 108 & (61) \\

\addlinespace\multicolumn{10}{l}{\emph{Additions to candidate}} \\
2a & \hladda{Finding/interpretation} & \textbf{51} & (44) & 72 & (57) & 61 & (50) & 54 & (46) \\
2b & \hladdb{Comparison} & 11 & (8) & 10 & (9) & 9 & (9) & \textbf{7} & (6) \\
2c & \hladdc{Ref. to prior report} & \textbf{0} & (0) & 1 & (1) & \textbf{0} & (0) & \textbf{0} & (0) \\
2d & \hladdd{Communication/followup} & 5 & (5) & 8 & (6) & 8 & (8) & \textbf{4} & (3) \\
2e & \hladde{Contradicting finding} & \textbf{0} & (0) & 1 & (1) & 3 & (3) & 1 & (1) \\
\midrule
\multicolumn{2}{l}{Total} & 67 & (49) & 92 & (63) & 81 & (58) & \textbf{66} & (48) \\

\addlinespace\multicolumn{10}{l}{\emph{Semantics of intersecting findings}} \\
3 & Incorrect location & \textbf{5} & (5) & 8 & (8) & 8 & (8) & 7 & (7) \\
4 & Incorrect severity & \textbf{6} & (6) & 7 & (7) & 7 & (7) & 9 & (9) \\

\addlinespace
5 & Other error & 31 & (23) & \textbf{30} & (23) & 33 & (29) & \textbf{30} & (21) \\
\bottomrule
\end{tabular}}
\hfill
\resizebox{\columnwidth}{!}{
\fbox{
\begin{sanseriffont}
\begin{tabular}{@{}p{\columnwidth}@{}}
\textbf{Reference:} Interval increase in vascular engorgement. No frank interstitial edema. \hlomma{No focal consolidations identified.} \\
\textbf{Candidate (M3):} interval increase in pulmonary vascular congestion without evidence of interstitial edema. \hladda{small right-sided pleural effusion.} \\

\addlinespace
\textbf{Reference:} Right lower lobe opacity, \hlomma{possibly atelectasis}, with associated moderate sized effusion. \\
\textbf{Candidate (M4):} \hladdb{persistent} right lower lobe opacity with associated effusion, \hladdb{mildly progressed from the preceding radiograph.} \\

\addlinespace
\textbf{Reference:} Multiloculated right pleural effusion unchanged \hlommc{since \_.} \hlomma{New linear and nodular opacities in the left upper lobe may represent carcinomatosis.} \hlommd{Findings were relayed to Dr. \_ by Dr. \_ \_ following review on \_ at approximiately 11:00 via telephone.} \\
\textbf{Candidate (M1):} stable appearance of multiple loculated right pleural effusion. \\

\addlinespace
\textbf{Reference:} Unchanged size and position of right-sided hydropneumothorax \hlommb{over the last \_-hour examination interval.} \\
\textbf{Candidate (M3):} \hladde{development of new right-sided hydropneumothorax} in this patient with \hladda{history of newly placed pigtail catheter.} \hladdd{referring physician, \_. \_ was paged at 4:45 p.m.} \\

\addlinespace
\textbf{Reference:} Little change in the severe bronchiectasis and \hlomma{emphysema}. \\
\textbf{Candidate (M3):} \hladde{unchanged} bibasilar bronchiectasis and bibasilar bronchiectasis.
\end{tabular}
\end{sanseriffont}
}}
\caption{Results of manual error analysis of 100 MIMIC-CXR reports. \ul{Left:} number of times each error occurred per method (percent of reports in gray, least errors per row in bold). \ul{Right:} example error annotations. Models: BertAbs (\textbf{M1}), GSum w/ Thresholding (\textbf{M2}), WGSum (\textbf{M3}), and WGSum+CL (\textbf{M4}) [best viewed in color].}
\label{tab:error-analysis-results}
\end{figure*}

\subsection{Results (RQ3)}
We report aggregated error counts and example annotations in \cref{tab:error-analysis-results}.\footnote{
To measure inter-annotator agreement (IAA), we calculate $F_1$ for span-annotations~\cite{Deleger:2012:AMIA} and Krippendorffs' Alpha for binary judgments~\cite{Krippendorff:1970:ALPHA}. Aggregated IAA: 1. Omissions: 0.61, 2. Additions: 0.60, 3. Incorrect Location: 0.25, and 4. Incorrect Severity: 0.41. IAA by error category for span-level annotations in~\cref{sec:appendix-iaa}.
}

Overall, we find that the prevalence of errors is comparable across the investigated methods, and that only 14--22\% of generated summaries are error-free.
The most common errors are omissions and additions of findings, which indicates that the models struggle to select relevant content (1a. 43--52\%; 2a. 44--57\%).
Compared with unguided summarization, there is a slight trend that guided methods reduce the risk of omissions, while only WGSum+CL succeeds at doing this without sacrificing precision.
Even though additions are common, they rarely contradict the reference (2e. 0--3\%).
Similarly, when both the reference and candidate present the same findings, errors related to their clinical correctness are rare (3. 5--8\%; 4. 6--9\%).

A surprising finding is the common omission and addition of clinicians' communications (1d. 16--20\%; 2d. 3--8\%).
By manual inspection (examples in \cref{tab:error-analysis-results}), we find that these are specific actions that a clinician performed after the examination such as informing colleagues about the findings, or recommending additional analysis.
Additions of this kind have likely no grounding in the underlying report.
To successfully generate such statements, models would require additional context information or guidance from a user.

\subsection{Discussion}
\label{sec:error-analysis-discussion}
Overall, our error analysis reveals that the key differences between model-generated impressions and radiologists' impressions relate to content selection (i.e., a tension between completeness/recall and relevance/precision).
We offer two hypotheses to explain the models' difficulties in this area.

First, there may be latent factors that explain which findings are included in the impression.
Among those factors could be patient demographics, the radiograph, prior exams and the clinical question.
Typically, this information is available to radiologists through the electronic health records, and is partly documented in the background section of radiology reports.
Early work explored using the background section as guidance~\cite{Zhang:2018:LOUHI}, but more recent work commonly excluded it in pre-processing~\cite{Sotudeh:2020:ACL,Hu:2021:ACL,Hu:2022:ACL}.
We present evaluation results when including the background and observe an overall improvement in almost all metrics for abstractive methods (\cref{sec:appendix-background-experiment}).
This improvement indicates that (i) additional context supports content selection, and (ii) it could be useful to explicitly model the background in guided summarization.

Second, we anecdotally observed a substantial degree of duplication in the MIMIC-CXR corpus, where reports with identical findings have different impressions (examples in \cref{sec:appendix-duplication-examples}).\footnote{11.9\% of the 122,500 MIMIC-CXR training reports have a findings section occurring more than once. Among those reports are only 1036 distinct impressions.}
This may lead to corpus-level inconsistencies preventing models to reliably learn the selection of findings.
We note that there can be numerous reasons for these duplication induced inconsistencies, including the presence of latent factors (see above) and remaining subjectivity/uncertainty in radiologists' assessments.
We leave the investigation of this aspect of data quality and potential effects of training data deduplication for future work.

\subsection{Limitations}
We note two limitations of this error analysis.

First, the analysis is based on comparing candidate impressions with reference impressions.
In the absence of the full clinical context, we argue that this is the most reliable benchmark for completeness and relevance of summaries.
However, we recognize that we cannot draw any conclusions about the factuality of additions with respect to the full report.
To give a first factuality estimate, we conducted a post-hoc analysis with RadNLI~\cite{Miura:2021:NAACL}.
Let $x_i$ be a sentence in report $\bm{x} = (x_1,\ldots,x_n)$, and $s$ be an addition span.
If RadNLI predicts a contradiction for any $(x_i, s)$ pair, we label this span as contradicting and neutral/entailed otherwise.
We find that between 23.4\% (BertAbs) and 29.3\% (GSum w/ Thresholding) of additions are contradicting, indicating that factuality is another challenge for current models (details in \cref{sec:appendix-radnli}).

Second, the sample size was driven by time and resource constraints ($N=100$).
To estimate representativeness of this sample, we compare descriptive statistics of the sample with those of the whole test set (length, novelty, compression), and observe that these largely agree (see \cref{sec:appendix-replication-error-analysis}).
While we believe that this sample is sufficient to support the qualitative conclusions about the failure modes of current methods, a larger study is warranted when the goal is to quantitatively compare the efficacy of different methods.

\section{Conclusion}
In this work, we revisited guided abstractive summarization of radiology reports.
We demonstrated that extractive summaries can be an effective guidance signal for the task, if we allow the length of this guidance signal to vary across reports, and thereby make the gap between domain-agnostic and domain-specific guidance smaller.
Furthermore, through a fine-grained error analysis of unguided and guided models we found that guidance successfully steers the content of summaries but that significant deficits in content selection persist.

We hope that this paper motivates future efforts on content selection mechanisms for radiology report summarization, their evaluation in other domains and languages, and on more comprehensive evaluation suites. We release our error annotations which can serve as a starting point for evaluating the efficacy of metrics in capturing these errors.

\section*{Ethical Considerations}
\paragraph{Privacy sensitive datasets.} Both the MIMIC-CXR dataset~\cite{Johnson:2019:SciData}, and the OpenI dataset~\cite{Demner-Fushman:2015:JAMIA} were fully de-identified by the dataset authors in compliance with applicable privacy laws (HIPAA).
This includes the removal of any protected health information that may directly or indirectly identify a patient.
Nevertheless, the data is still privacy sensitive, and special care was taken to only process it within secured computing infrastructure.

\paragraph{Intended use.} We believe that the proposed methods can improve the workflow of clinicians both by reducing the documentation effort and encouraging higher-quality reporting, and thereby improving patient care.
However, as our results and discussion show, state-of-the-art summarization methods may not have the desired level of quality that is needed in high-stakes domains such as the clinical context.
Therefore, our work is not to be understood in the context of a system that can be deployed, but rather as a step toward a better understanding of the shortcomings of current text summarization methods and providing insight into how these can solved.

\section*{Supplementary Materials Availability Statement}

\begin{itemize}[noitemsep]
\item Detailed analysis, hyperparameters, and annotation guidelines are available in~\cref{sec:appendix-analysis,sec:appendix-replication-models,sec:appendix-replication-error-analysis,sec:appendix-annotation-guidelines}.
\item Source code to reproduce all experiments is available from \rurl{github.com/jantrienes/inlg2023-radsum/}
\item The expert annotations of summarization errors are available from \rurl{github.com/jantrienes/inlg2023-radsum/} under the \href{https://physionet.org/content/mimic-cxr/view-license/2.0.0/}{PhysioNet Credentialed Health Data License 1.5.0}.
\item The MIMIC-CXR (v2.0.0) dataset is available from \rurl{physionet.org/content/mimic-cxr/} under the \href{https://physionet.org/content/mimic-cxr/view-license/2.0.0/}{PhysioNet Credentialed Health Data License 1.5.0}.
\item The OpenI dataset is available from \rurl{openi.nlm.nih.gov} (no license terms stated)
\end{itemize}

\section*{Acknowledgements}
We thank three anonymous reviewers for their insightful comments, Dennis Aumiller for giving feedback on an earlier version of this work, and Jinpeng Hu for helping with the reproduction of WGSum+CL. We also thank our medical students for valuable discussions and their annotation effort.

\bibliography{bibliography}

\begin{thebibliography}{46}
\expandafter\ifx\csname natexlab\endcsname\relax\def\natexlab#1{#1}\fi

\bibitem[{Cai et~al.(2021)Cai, Liu, Han, Yang, Liu, and
  Liu}]{Cai:2021:TransMultimedia}
Xiaoyan Cai, Sen Liu, Junwei Han, Libin Yang, Zhenguo Liu, and Tianming Liu.
  2021.
\newblock \href {https://doi.org/10.1109/TMM.2021.3132724} {{ChestXRayBERT}:
  {A} pretrained language model for chest radiology report summarization}.
\newblock \emph{IEEE Transactions on Multimedia}, 25:845--855.

\bibitem[{Cao et~al.(2018)Cao, Li, Li, and Wei}]{Cao:2018:ACL}
Ziqiang Cao, Wenjie Li, Sujian Li, and Furu Wei. 2018.
\newblock \href {https://doi.org/10.18653/v1/P18-1015} {Retrieve, rerank and
  rewrite: {S}oft template based neural summarization}.
\newblock In \emph{Proceedings of the 56th Annual Meeting of the Association
  for Computational Linguistics (ACL)}, pages 152--161.

\bibitem[{Chen and Bansal(2018)}]{Chen:2018:ACL}
Yen{-}Chun Chen and Mohit Bansal. 2018.
\newblock \href {https://doi.org/10.18653/v1/P18-1063} {Fast abstractive
  summarization with reinforce-selected sentence rewriting}.
\newblock In \emph{Proceedings of the 56th Annual Meeting of the Association
  for Computational Linguistics (ACL)}, pages 675--686.

\bibitem[{Delbrouck et~al.(2022)Delbrouck, Varma, and
  Langlotz}]{Delbrouck:2022:ML4H}
Jean-Benoit Delbrouck, Maya Varma, and Curtis~P. Langlotz. 2022.
\newblock \href {https://doi.org/10.48550/arXiv.2211.08584} {Toward expanding
  the scope of radiology report summarization to multiple anatomies and
  modalities}.
\newblock In \emph{Proceedings of the 2nd Machine Learning for Health symposium
  (ML4H), Extended Abstract Collection}.

\bibitem[{Deleger et~al.(2012)Deleger, Li, Lingren, Kaiser, Molnar,
  Stoutenborough, Kouril, Marsolo, and Solti}]{Deleger:2012:AMIA}
Louise Deleger, Qi~Li, Todd Lingren, Megan Kaiser, Katalin Molnar, Laura
  Stoutenborough, Michal Kouril, Keith Marsolo, and Imre Solti. 2012.
\newblock \href {https://www.ncbi.nlm.nih.gov/pmc/articles/PMC3540456/}
  {Building gold standard corpora for medical natural language processing
  tasks}.
\newblock In \emph{AMIA Annual Symposium Proceedings}, pages 144--153.

\bibitem[{Demner-Fushman et~al.(2015)Demner-Fushman, Kohli, Rosenman, Shooshan,
  Rodriguez, Antani, Thoma, and McDonald}]{Demner-Fushman:2015:JAMIA}
Dina Demner-Fushman, Marc~D. Kohli, Marc~B. Rosenman, Sonya~E. Shooshan,
  Laritza Rodriguez, Sameer Antani, George~R. Thoma, and Clement~J. McDonald.
  2015.
\newblock \href {https://doi.org/10.1093/jamia/ocv080} {Preparing a collection
  of radiology examinations for distribution and retrieval}.
\newblock \emph{Journal of the American Medical Informatics Association},
  23(2):304--310.

\bibitem[{Devlin et~al.(2019)Devlin, Chang, Lee, and
  Toutanova}]{Devlin:2019:NAACL}
Jacob Devlin, Ming-Wei Chang, Kenton Lee, and Kristina Toutanova. 2019.
\newblock \href {https://doi.org/10.18653/v1/N19-1423} {{BERT}: {P}re-training
  of deep bidirectional transformers for language understanding}.
\newblock In \emph{Proceedings of the 2019 Conference of the North American
  Chapter of the Association for Computational Linguistics: Human Language
  Technologies (NAACL)}, pages 4171--4186.

\bibitem[{Dou et~al.(2021)Dou, Liu, Hayashi, Jiang, and
  Neubig}]{Dou:2021:NAACL}
Zi{-}Yi Dou, Pengfei Liu, Hiroaki Hayashi, Zhengbao Jiang, and Graham Neubig.
  2021.
\newblock \href {https://doi.org/10.18653/v1/2021.naacl-main.384} {{GSum}: {A}
  general framework for guided neural abstractive summarization}.
\newblock In \emph{Proceedings of the 2021 Conference of the North American
  Chapter of the Association for Computational Linguistics: Human Language
  Technologies (NAACL)}, pages 4830--4842.

\bibitem[{Fabbri et~al.(2021)Fabbri, Kry{\'s}ci{\'n}ski, McCann, Xiong, Socher,
  and Radev}]{Fabbri:2021:TACL}
Alexander~R. Fabbri, Wojciech Kry{\'s}ci{\'n}ski, Bryan McCann, Caiming Xiong,
  Richard Socher, and Dragomir Radev. 2021.
\newblock \href {https://doi.org/10.1162/tacl_a_00373} {{S}umm{E}val:
  {R}e-evaluating summarization evaluation}.
\newblock \emph{Transactions of the Association for Computational Linguistics},
  9:391--409.

\bibitem[{Fan et~al.(2018)Fan, Grangier, and Auli}]{Fan:2018:ACLws}
Angela Fan, David Grangier, and Michael Auli. 2018.
\newblock \href {https://doi.org/10.18653/v1/W18-2706} {Controllable
  abstractive summarization}.
\newblock In \emph{Proceedings of the 2nd Workshop on Neural Machine
  Translation and Generation}, pages 45--54.

\bibitem[{Gershanik et~al.(2011)Gershanik, Lacson, and
  Khorasani}]{Gershanik:2011:AMIA}
Esteban~F. Gershanik, Ronilda Lacson, and Ramin Khorasani. 2011.
\newblock \href {https://ncbi.nlm.nih.gov/pmc/articles/PMC3243237} {Critical
  finding capture in the impression section of radiology reports}.
\newblock In \emph{AMIA Annual Symposium Proceedings}, pages 465--469.

\bibitem[{He et~al.(2022)He, Kry{\'s}ci{\'n}ski, McCann, Rajani, and
  Xiong}]{He:2022:EMNLP}
Junxian He, Wojciech Kry{\'s}ci{\'n}ski, Bryan McCann, Nazneen Rajani, and
  Caiming Xiong. 2022.
\newblock \href {https://aclanthology.org/2022.emnlp-main.396} {{CTRL}sum:
  Towards generic controllable text summarization}.
\newblock In \emph{Proceedings of the 2022 Conference on Empirical Methods in
  Natural Language Processing (EMNLP)}, pages 5879--5915.

\bibitem[{Hu et~al.(2021)Hu, Li, Chen, Shen, Song, Wan, and
  Chang}]{Hu:2021:ACL}
Jinpeng Hu, Jianling Li, Zhihong Chen, Yaling Shen, Yan Song, Xiang Wan, and
  Tsung-Hui Chang. 2021.
\newblock \href {https://doi.org/10.18653/v1/2021.findings-acl.441} {Word graph
  guided summarization for radiology findings}.
\newblock In \emph{Findings of the Association for Computational Linguistics
  (ACL-IJCNLP)}, pages 4980--4990.

\bibitem[{Hu et~al.(2022)Hu, Li, Chen, Li, Wan, and Chang}]{Hu:2022:ACL}
Jinpeng Hu, Zhuo Li, Zhihong Chen, Zhen Li, Xiang Wan, and Tsung-Hui Chang.
  2022.
\newblock \href {https://doi.org/10.18653/v1/2022.acl-long.320} {Graph enhanced
  contrastive learning for radiology findings summarization}.
\newblock In \emph{Proceedings of the 60th Annual Meeting of the Association
  for Computational Linguistics (ACL)}, pages 4677--4688.

\bibitem[{Huang et~al.(2020)Huang, Cui, Yang, Bao, Wang, Xie, and
  Zhang}]{Huang:2020:EMNLP}
Dandan Huang, Leyang Cui, Sen Yang, Guangsheng Bao, Kun Wang, Jun Xie, and Yue
  Zhang. 2020.
\newblock \href {https://doi.org/10.18653/v1/2020.emnlp-main.33} {What have we
  achieved on text summarization?}
\newblock In \emph{Proceedings of the 2020 Conference on Empirical Methods in
  Natural Language Processing (EMNLP)}, pages 446--469.

\bibitem[{Irvin et~al.(2019)Irvin, Rajpurkar, Ko, Yu, Ciurea-Ilcus, Chute,
  Marklund, Haghgoo, Ball, Shpanskaya, Seekins, Mong, Halabi, Sandberg, Jones,
  Larson, Langlotz, Patel, Lungren, and Ng}]{Irvin:2019:AAAI}
Jeremy Irvin, Pranav Rajpurkar, Michael Ko, Yifan Yu, Silviana Ciurea-Ilcus,
  Chris Chute, Henrik Marklund, Behzad Haghgoo, Robyn Ball, Katie Shpanskaya,
  Jayne Seekins, David~A. Mong, Safwan~S. Halabi, Jesse~K. Sandberg, Ricky
  Jones, David~B. Larson, Curtis~P. Langlotz, Bhavik~N. Patel, Matthew~P.
  Lungren, and Andrew~Y. Ng. 2019.
\newblock \href {https://doi.org/10.1609/aaai.v33i01.3301590} {{CheXpert}: {A}
  large chest radiograph dataset with uncertainty labels and expert
  comparison}.
\newblock In \emph{Proceedings of the AAAI Conference on Artificial
  Intelligence}, pages 590--597.

\bibitem[{Jia et~al.(2021)Jia, Cao, Shi, Fang, Yin, and Wang}]{Jia:2021:AAAI}
Ruipeng Jia, Yanan Cao, Haichao Shi, Fang Fang, Pengfei Yin, and Shi Wang.
  2021.
\newblock \href {https://ojs.aaai.org/index.php/AAAI/article/view/17552}
  {Flexible non-autoregressive extractive summarization with threshold: {H}ow
  to extract a non-fixed number of summary sentences}.
\newblock In \emph{Proceedings of the AAAI Conference on Artificial
  Intelligence}, pages 13134--13142.

\bibitem[{Johnson et~al.(2019{\natexlab{a}})Johnson, Pollard, Berkowitz,
  Greenbaum, Lungren, Deng, Mark, and Horng}]{Johnson:2019:SciData}
Alistair E.~W. Johnson, Tom~J. Pollard, Seth~J. Berkowitz, Nathaniel~R.
  Greenbaum, Matthew~P. Lungren, Chih-ying Deng, Roger~G. Mark, and Steven
  Horng. 2019{\natexlab{a}}.
\newblock \href {https://doi.org/10.1038/s41597-019-0322-0} {{MIMIC-CXR}, a
  de-identified publicly available database of chest radiographs with free-text
  reports}.
\newblock \emph{Scientific Data}, 6(317).

\bibitem[{Johnson et~al.(2019{\natexlab{b}})Johnson, Pollard, Greenbaum,
  Lungren, ying Deng, Peng, Lu, Mark, Berkowitz, and
  Horng}]{Johnson:2019:arXiv}
Alistair E.~W. Johnson, Tom~J. Pollard, Nathaniel~R. Greenbaum, Matthew~P.
  Lungren, Chih ying Deng, Yifan Peng, Zhiyong Lu, Roger~G. Mark, Seth~J.
  Berkowitz, and Steven Horng. 2019{\natexlab{b}}.
\newblock \href {https://arxiv.org/abs/1901.07042} {{MIMIC-CXR-JPG}, a large
  publicly available database of labeled chest radiographs}.
\newblock \emph{CoRR}, abs/1901.07042.

\bibitem[{Kahn et~al.(2009)Kahn, Langlotz, Burnside, Carrino, Channin,
  Hovsepian, and Rubin}]{Kahn:2009:Radiology}
Charles~E. Kahn, Curtis~P. Langlotz, Elizabeth~S. Burnside, John~A. Carrino,
  David~S. Channin, David~M. Hovsepian, and Daniel~L. Rubin. 2009.
\newblock \href {https://doi.org/10.1148/radiol.2523081992} {Toward best
  practices in radiology reporting}.
\newblock \emph{Radiology}, 252(3):852--856.

\bibitem[{Karn et~al.(2022)Karn, Liu, Schütze, and Farri}]{Karn:2022:ACL}
Sanjeev~Kumar Karn, Ning Liu, Hinrich Schütze, and Oladimeji Farri. 2022.
\newblock \href {https://doi.org/10.18653/v1/2022.acl-long.109} {Differentiable
  multi-agent actor-critic for multi-step radiology report summarization}.
\newblock In \emph{Proceedings of the 60th Annual Meeting of the Association
  for Computational Linguistics (ACL)}, pages 1542--1553.

\bibitem[{Krippendorff(1970)}]{Krippendorff:1970:ALPHA}
Klaus Krippendorff. 1970.
\newblock Bivariate agreement coefficients for reliability of data.
\newblock \emph{Sociological methodology}, 2:139--150.

\bibitem[{Kry{\'s}ci{\'n}ski et~al.(2020)Kry{\'s}ci{\'n}ski, McCann, Xiong, and
  Socher}]{Kryscinski:2020:EMNLP}
Wojciech Kry{\'s}ci{\'n}ski, Bryan McCann, Caiming Xiong, and Richard Socher.
  2020.
\newblock \href {https://doi.org/10.18653/v1/2020.emnlp-main.750} {Evaluating
  the factual consistency of abstractive text summarization}.
\newblock In \emph{Proceedings of the 2020 Conference on Empirical Methods in
  Natural Language Processing (EMNLP)}, pages 9332--9346.

\bibitem[{Li et~al.(2018)Li, Xu, Li, and Gao}]{Li:2018:NAACL}
Chenliang Li, Weiran Xu, Si~Li, and Sheng Gao. 2018.
\newblock \href {https://doi.org/10.18653/v1/n18-2009} {Guiding generation for
  abstractive text summarization based on key information guide network}.
\newblock In \emph{Proceedings of the 2018 Conference of the North American
  Chapter of the Association for Computational Linguistics: Human Language
  Technologies (NAACL)}, pages 55--60.

\bibitem[{Lin(2004)}]{Lin:2004:WS}
Chin-Yew Lin. 2004.
\newblock \href {https://aclanthology.org/W04-1013/} {{ROUGE}: {A} package for
  automatic evaluation of summaries}.
\newblock In \emph{Text Summarization Branches Out}, pages 74--81.

\bibitem[{Liu and Lapata(2019)}]{Liu:2019:EMNLP}
Yang Liu and Mirella Lapata. 2019.
\newblock \href {https://doi.org/10.18653/v1/D19-1387} {Text summarization with
  pretrained encoders}.
\newblock In \emph{Proceedings of the 2019 Conference on Empirical Methods in
  Natural Language Processing and the 9th International Joint Conference on
  Natural Language Processing (EMNLP-IJCNLP)}, pages 3730--3740.

\bibitem[{Lommel et~al.(2014)Lommel, Uszkoreit, and
  Burchardt}]{Lommel:2014:MQM}
Arle Lommel, Hans Uszkoreit, and Aljoscha Burchardt. 2014.
\newblock \href {https://doi.org/10.5565/rev/tradumatica.77} {Multidimensional
  quality metrics {(MQM)}: {A} framework for declaring and describing
  translation quality metrics}.
\newblock \emph{Tradum{\`a}tica}, 2014(12):455--463.

\bibitem[{MacAvaney et~al.(2019)MacAvaney, Sotudeh, Cohan, Goharian, Talati,
  and Filice}]{MacAvaney:2019:SIGIR}
Sean MacAvaney, Sajad Sotudeh, Arman Cohan, Nazli Goharian, Ish Talati, and
  Ross~W. Filice. 2019.
\newblock \href {https://doi.org/10.1145/3331184.3331319} {Ontology-aware
  clinical abstractive summarization}.
\newblock In \emph{Proceedings of the 42nd International {ACM} {SIGIR}
  Conference on Research and Development in Information Retrieval (SIGIR)},
  pages 1013--1016.

\bibitem[{Maynez et~al.(2020)Maynez, Narayan, Bohnet, and
  McDonald}]{Maynez:2020:ACL}
Joshua Maynez, Shashi Narayan, Bernd Bohnet, and Ryan McDonald. 2020.
\newblock \href {https://doi.org/10.18653/v1/2020.acl-main.173} {On
  faithfulness and factuality in abstractive summarization}.
\newblock In \emph{Proceedings of the 58th Annual Meeting of the Association
  for Computational Linguistics (ACL)}, pages 1906--1919.

\bibitem[{Miura et~al.(2021)Miura, Zhang, Tsai, Langlotz, and
  Jurafsky}]{Miura:2021:NAACL}
Yasuhide Miura, Yuhao Zhang, Emily Tsai, Curtis Langlotz, and Dan Jurafsky.
  2021.
\newblock \href {https://doi.org/10.18653/v1/2021.naacl-main.416} {Improving
  factual completeness and consistency of image-to-text radiology report
  generation}.
\newblock In \emph{Proceedings of the 2021 Conference of the North American
  Chapter of the Association for Computational Linguistics (NAACL)}, pages
  5288--5304.

\bibitem[{Nallapati et~al.(2017)Nallapati, Zhai, and
  Zhou}]{Nallapati:2017:AAAI}
Ramesh Nallapati, Feifei Zhai, and Bowen Zhou. 2017.
\newblock \href
  {https://aaai.org/ocs/index.php/AAAI/AAAI17/paper/view/14636/14080}
  {{SummaRuNNer}: {A} recurrent neural network based sequence model for
  extractive summarization of documents}.
\newblock In \emph{Proceedings of the AAAI Conference on Artificial
  Intelligence}, pages 3075--3081.

\bibitem[{Nallapati et~al.(2016)Nallapati, Zhou, dos Santos, \c{C}a\u{g}lar
  G\.{u}l\c{c}ehre, and Xiang}]{Nallapati:2016:CoNLL}
Ramesh Nallapati, Bowen Zhou, Cicero dos Santos, \c{C}a\u{g}lar
  G\.{u}l\c{c}ehre, and Bing Xiang. 2016.
\newblock \href {https://doi.org/10.18653/v1/k16-1028} {Abstractive text
  summarization using sequence-to-sequence {RNNs} and beyond}.
\newblock In \emph{Proceedings of the 20th {SIGNLL} Conference on Computational
  Natural Language Learning (CoNLL)}, pages 280--290.

\bibitem[{Narayan et~al.(2021)Narayan, Zhao, Maynez, Sim{\~o}es, Nikolaev, and
  McDonald}]{Narayan:2021:TACL}
Shashi Narayan, Yao Zhao, Joshua Maynez, Gon{\c{c}}alo Sim{\~o}es, Vitaly
  Nikolaev, and Ryan McDonald. 2021.
\newblock \href {https://doi.org/10.1162/tacl_a_00438} {Planning with learned
  entity prompts for abstractive summarization}.
\newblock \emph{Transactions of the Association for Computational Linguistics},
  9:1475--1492.

\bibitem[{Rush et~al.(2015)Rush, Chopra, and Weston}]{Rush:2015:EMNLP}
Alexander~M. Rush, Sumit Chopra, and Jason Weston. 2015.
\newblock \href {https://doi.org/10.18653/v1/D15-1044} {A neural attention
  model for abstractive sentence summarization}.
\newblock In \emph{Proceedings of the 2015 Conference on Empirical Methods in
  Natural Language Processing (EMNLP)}, pages 379--389.

\bibitem[{Sanh et~al.(2019)Sanh, Debut, Chaumond, and
  Wolf}]{Sanh:2019:DistilBERT}
Victor Sanh, Lysandre Debut, Julien Chaumond, and Thomas Wolf. 2019.
\newblock \href {http://arxiv.org/abs/1910.01108} {{DistilBERT}, a distilled
  version of {BERT:} smaller, faster, cheaper and lighter}.
\newblock In \emph{Proceedings of the 5th Workshop on Energy Efficient Machine
  Learning and Cognitive Computing (EMC2-NIPS)}.

\bibitem[{Sotudeh et~al.(2020)Sotudeh, Goharian, and Filice}]{Sotudeh:2020:ACL}
Sajad Sotudeh, Nazli Goharian, and Ross~W. Filice. 2020.
\newblock \href {https://doi.org/10.18653/v1/2020.acl-main.172} {Attend to
  medical ontologies: {C}ontent selection for clinical abstractive
  summarization}.
\newblock In \emph{Proceedings of the 58th Annual Meeting of the Association
  for Computational Linguistics (ACL)}, pages 1899--1905.

\bibitem[{van Miltenburg et~al.(2021)van Miltenburg, Clinciu, Du{\v{s}}ek,
  Gkatzia, Inglis, Lepp{\"a}nen, Mahamood, Manning, Schoch, Thomson, and
  Wen}]{Van-Miltenburg:2021:INLG}
Emiel van Miltenburg, Miruna Clinciu, Ond{\v{r}}ej Du{\v{s}}ek, Dimitra
  Gkatzia, Stephanie Inglis, Leo Lepp{\"a}nen, Saad Mahamood, Emma Manning,
  Stephanie Schoch, Craig Thomson, and Luou Wen. 2021.
\newblock \href {https://aclanthology.org/2021.inlg-1.14} {Underreporting of
  errors in {NLG} output, and what to do about it}.
\newblock In \emph{Proceedings of the 14th International Conference on Natural
  Language Generation (INLG)}, pages 140--153.

\bibitem[{Van~Veen et~al.(2023)Van~Veen, Van~Uden, Attias, Pareek, Bluethgen,
  Polacin, Chiu, Delbrouck, Zambrano~Chaves, Langlotz, Chaudhari, and
  Pauly}]{VanVeen:2023:ACL-ws}
Dave Van~Veen, Cara Van~Uden, Maayane Attias, Anuj Pareek, Christian Bluethgen,
  Malgorzata Polacin, Wah Chiu, Jean-Benoit Delbrouck, Juan Zambrano~Chaves,
  Curtis Langlotz, Akshay Chaudhari, and John Pauly. 2023.
\newblock \href {https://aclanthology.org/2023.bionlp-1.42} {{R}ad{A}dapt:
  {R}adiology report summarization via lightweight domain adaptation of large
  language models}.
\newblock In \emph{The 22nd Workshop on Biomedical Natural Language Processing
  and BioNLP Shared Tasks}, pages 449--460.

\bibitem[{Vaswani et~al.(2017)Vaswani, Shazeer, Parmar, Uszkoreit, Jones,
  Gomez, Kaiser, and Polosukhin}]{Vaswani:2017:NIPS}
Ashish Vaswani, Noam Shazeer, Niki Parmar, Jakob Uszkoreit, Llion Jones,
  Aidan~N. Gomez, {\L}ukasz Kaiser, and Illia Polosukhin. 2017.
\newblock \href
  {https://proceedings.neurips.cc/paper/2017/hash/3f5ee243547dee91fbd053c1c4a845aa-Abstract.html}
  {Attention is all you need}.
\newblock In \emph{Advances in Neural Information Processing Systems
  (NeurIPS)}, pages 5998--6008.

\bibitem[{Xie et~al.(2023)Xie, Zhou, Peng, and Wang}]{Xie:2023:arXiv}
Qianqian Xie, Jiayu Zhou, Yifan Peng, and Fei Wang. 2023.
\newblock \href {https://arxiv.org/abs/2303.08335} {{FactReranker}:
  {F}act-guided reranker for faithful radiology report summarization}.
\newblock \emph{CoRR}, abs/2303.08335.

\bibitem[{Yu et~al.(2022)Yu, Endo, Krishnan, Pan, Tsai, Reis, Fonseca, Lee,
  Abad, Ng, Langlotz, Venugopal, and Rajpurkar}]{Yu:2022:medRxiv}
Feiyang Yu, Mark Endo, Rayan Krishnan, Ian Pan, Andy Tsai, Eduardo~Pontes Reis,
  Eduardo Kaiser Ururahy~Nunes Fonseca, Henrique Min~Ho Lee, Zahra
  Shakeri~Hossein Abad, Andrew~Y. Ng, Curtis~P. Langlotz, Vasantha~Kumar
  Venugopal, and Pranav Rajpurkar. 2022.
\newblock \href {https://doi.org/10.1101/2022.08.30.22279318} {Evaluating
  progress in automatic chest x-ray radiology report generation}.
\newblock \emph{medRxiv}.

\bibitem[{Zhang et~al.(2020{\natexlab{a}})Zhang, Kishore, Wu, Weinberger, and
  Artzi}]{Zhang:2020:ICLR}
Tianyi Zhang, Varsha Kishore, Felix Wu, Kilian~Q. Weinberger, and Yoav Artzi.
  2020{\natexlab{a}}.
\newblock \href {https://openreview.net/forum?id=SkeHuCVFDr} {{BERTScore}:
  {E}valuating text generation with {BERT}}.
\newblock In \emph{Proceedings of the 8th International Conference on Learning
  Representations (ICLR)}.

\bibitem[{Zhang et~al.(2018)Zhang, Ding, Qian, Manning, and
  Langlotz}]{Zhang:2018:LOUHI}
Yuhao Zhang, Daisy~Yi Ding, Tianpei Qian, Christopher~D. Manning, and Curtis~P.
  Langlotz. 2018.
\newblock \href {https://doi.org/10.18653/v1/w18-5623} {Learning to summarize
  radiology findings}.
\newblock In \emph{Proceedings of the 9th International Workshop on Health Text
  Mining and Information Analysis (LOUHI)}, pages 204--213.

\bibitem[{Zhang et~al.(2020{\natexlab{b}})Zhang, Merck, Tsai, Manning, and
  Langlotz}]{Zhang:2020:ACL}
Yuhao Zhang, Derek Merck, Emily~Bao Tsai, Christopher~D. Manning, and Curtis~P.
  Langlotz. 2020{\natexlab{b}}.
\newblock \href {https://doi.org/10.18653/v1/2020.acl-main.458} {Optimizing the
  factual correctness of a summary: {A} study of summarizing radiology
  reports}.
\newblock In \emph{Proceedings of the 58th Annual Meeting of the Association
  for Computational Linguistics (ACL)}, pages 5108--5120.

\bibitem[{Zhang et~al.(2021)Zhang, Zhang, Qi, Manning, and
  Langlotz}]{Zhang:2021:JAMIA}
Yuhao Zhang, Yuhui Zhang, Peng Qi, Christopher~D. Manning, and Curtis~P.
  Langlotz. 2021.
\newblock \href {https://doi.org/10.1093/jamia/ocab090} {Biomedical and
  clinical english model packages for the stanza python {NLP} library}.
\newblock \emph{Journal of the American Medical Informatics Association},
  28(9):1892--1899.

\bibitem[{Zhu et~al.(2023)Zhu, Yang, Wu, and Zhang}]{Zhu:2023:arXiv}
Yunqi Zhu, Xuebing Yang, Yuanyuan Wu, and Wensheng Zhang. 2023.
\newblock \href {https://arxiv.org/abs/2302.04001} {Leveraging summary guidance
  on medical report summarization}.
\newblock \emph{CoRR}, abs/2302.04001.

\end{thebibliography}
\bibliographystyle{acl_natbib}

\appendix
\section{Detailed Analysis}
\label{sec:appendix-analysis}
To support replication, this section provides supplementary analysis on the results of the main part.

\subsection{Target Impression Length Distribution and Evaluation by Target Length}
\label{sec:appendix-target-summary-length}
\begin{figure*}[t]
\centering
\includegraphics[width=\textwidth]{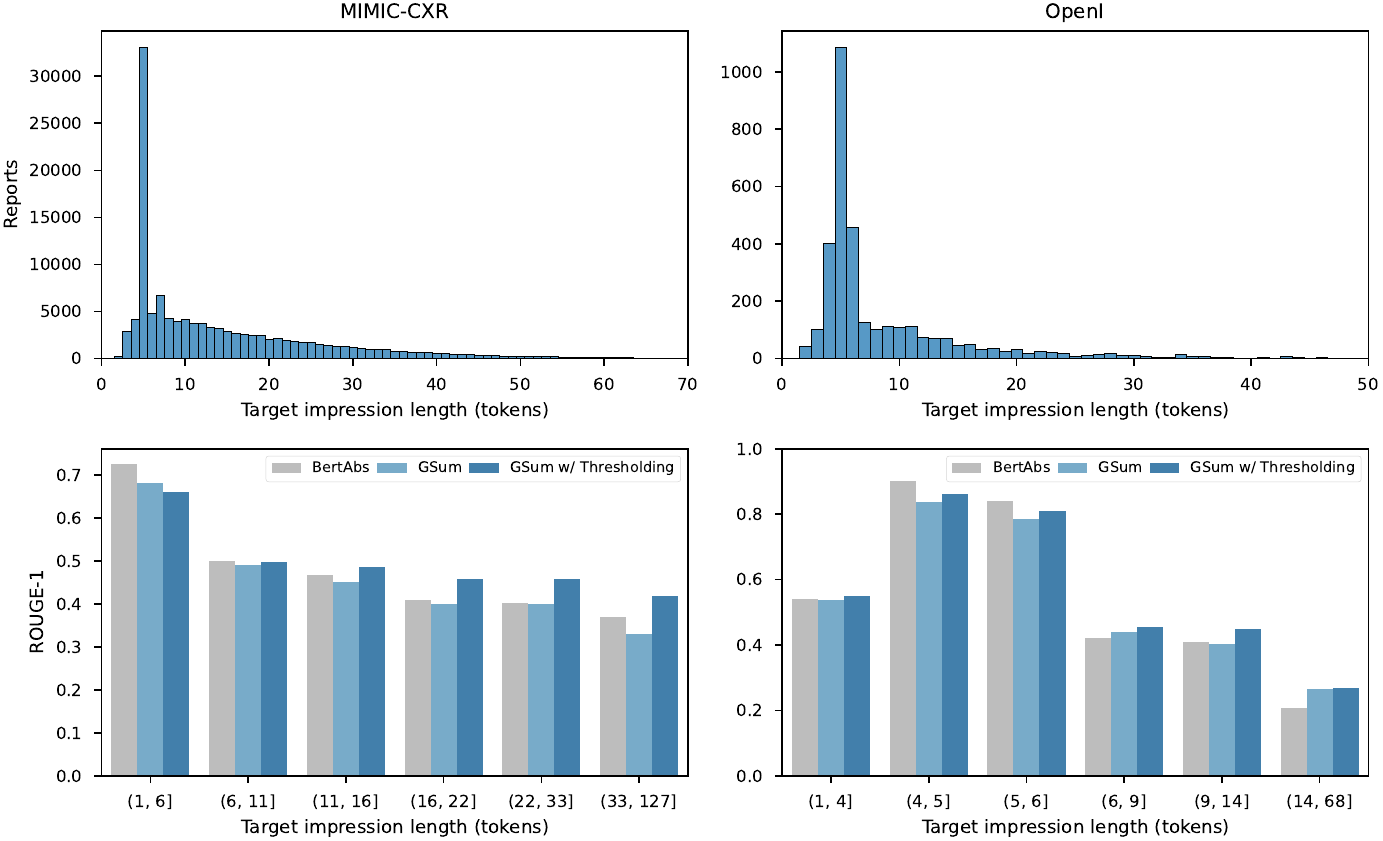}
\caption{\ul{Top row:} distribution of target impression lengths. \ul{Bottom row:} ROUGE-1 by target length for BertAbs (unguided summarization), GSum (fixed-length guidance) and GSum w/ Thresholding (variable-length guidance).}
\label{fig:target-summary-length}
\end{figure*}

We demonstrated in the main part that variable-length guidance helps to adapt to varying target lengths.
To better interpret this result, we plot the length distribution of target summaries and the ROUGE-1 score by target-length interval in \cref{fig:target-summary-length}.
It can be observed that the length distribution has a long tail with a peak around 4-5 tokens.
Impressions of this length are standard phrasings to indicate that no abnormalities were found (e.g., \emph{``No evidence of acute findings''}).

\subsection{Evaluating GSum in an Oracle Setting}
\label{sec:appendix-gsum-oracle-experiment}
As a supplement to the oracle experiment in \cref{sec:results-gsum-fixed}, we provide all metrics for the three inference settings of GSum in \cref{tab:appendix-gsum-oracle}: (i) automatic fixed-length guidance (i.e., extracted from BertExt with $k=1$), (ii) automatic variable-length guidance but with an oracle length (i.e., BertExt with $k = |\text{OracleExt}(\bm{x}, \bm{y})|$), and (iii) oracle guidance (i.e., $\bm{g} = \text{OracleExt}(\bm{x}, \bm{y})$).

\begin{table}[t]
\small
\centering
\setlength\tabcolsep{2pt}
\begin{tabular}{lrrrrr}
\toprule
\textbf{MIMIC-CXR} & \textbf{R-1} & \textbf{R-2} & \textbf{R-L} & \textbf{BS} & \textbf{Fact.} \\
\midrule
\emph{Guidance signal for GSum}\\
Fixed~\cite{Dou:2021:NAACL} & 46.3 & 32.7 & 44.7 & 57.4 & 46.6 \\
Oracle Length & 51.7 & 36.3 & 49.6 & 61.2 & 52.4 \\
Oracle Length + Content & 58.5 & 42.0 & 56.2 & 66.0 & 60.0 \\
\midrule
\textbf{OpenI} & & & & & \\
\midrule
\emph{Guidance signal for GSum}\\
Fixed~\cite{Dou:2021:NAACL} & 60.1 & 49.6 & 59.8 & 67.0 & 40.0 \\
Oracle Length & 63.9 & 53.0 & 63.5 & 69.4 & 42.3 \\
Oracle Length + Content & 68.8 & 56.7 & 68.3 & 72.7 & 45.1 \\
\bottomrule
\end{tabular}
\caption{Evaluating GSum in an oracle setting. \emph{Fixed} is reproduced from~\cref{tab:results}.}
\label{tab:appendix-gsum-oracle}
\end{table}

\subsection{BertExt: Evaluating Fixed-length Settings}
\label{sec:appendx-fixed-k-testing}
To evaluate if larger values for $k$ in the fixed-summary length setting would improve the effectiveness of BertExt, we generate summaries for all settings of $k = \{1,...,5\}$.
Analogously, we provide these summaries as guidance signal to GSum.
\cref{tab:appendix-fixed-k-testing} reports the results of this experiment.
While we find that larger settings of $k$ lead to an increase in recall, we see an equally strong drop in precision, both on BertExt and GSum which demonstrates the necessity of variable-length extractive guidance.

\def\SecBertExtStatic{\multicolumn{11}{l}{BertExt with fixed-length summaries} \\\midrule}
\def\SecGsumStatic{\midrule \multicolumn{11}{l}{GSum with fixed-length guidance extracted from BertExt} \\\midrule}
\begin{table}[t]
\small
\centering
\setlength\tabcolsep{3pt}
\resizebox{\linewidth}{!}{
\begin{tabular}{lrrrrrrrrrr}
\toprule
{} & \multicolumn{5}{c}{\textbf{MIMIC-CXR}} & \multicolumn{5}{c}{\textbf{OpenI}} \\
\cmidrule(lr){2-6}
\cmidrule(lr){7-11}
{} & R-1 & R-2 & R-L & B\textsubscript{P} & B\textsubscript{R} & R-1 & R-2 & R-L & B\textsubscript{P} & B\textsubscript{R} \\
\midrule
\SecBertExtStatic
$k=1$ & 32.7 & 18.1 & 30.0 & \textbf{45.2} & 40.1 & \textbf{23.6} & \textbf{7.4} & \textbf{22.6} & \textbf{33.6} & 32.3 \\
$k=2$ & \textbf{34.1} & \textbf{18.6} & \textbf{31.3} & 40.9 & 50.1 & 19.7 & 6.7 & 18.9 & 28.3 & 39.9 \\
$k=3$ & 31.7 & 17.0 & 29.2 & 37.0 & 53.5 & 17.4 & 6.1 & 16.6 & 25.9 & 42.8 \\
$k=4$ & 29.1 & 15.4 & 26.8 & 34.0 & 54.6 & 15.8 & 5.5 & 15.1 & 24.0 & 43.7 \\
$k=5$ & 27.2 & 14.3 & 25.2 & 32.2 & \textbf{54.9} & 15.1 & 5.2 & 14.4 & 23.3 & \textbf{44.1} \\
\SecGsumStatic
$k=1$ & \textbf{46.3} & \textbf{32.7} & \textbf{44.7} & \textbf{64.6} & 52.8 & \textbf{60.1} & \textbf{49.6} & \textbf{59.8} & \textbf{67.0} & \textbf{68.5} \\
$k=2$ & \textbf{46.3} & 30.3 & 44.2 & 58.1 & 58.5 & 54.3 & 43.2 & 53.9 & 61.2 & 66.2 \\
$k=3$ & 44.1 & 27.7 & 41.9 & 53.6 & 59.9 & 54.6 & 43.2 & 54.1 & 61.6 & 67.3 \\
$k=4$ & 42.2 & 26.0 & 40.2 & 50.4 & \textbf{60.2} & 53.5 & 42.1 & 53.1 & 60.1 & 67.5 \\
$k=5$ & 40.8 & 24.6 & 38.8 & 48.3 & 60.1 & 52.7 & 41.3 & 52.2 & 59.5 & 67.5 \\
\bottomrule
\end{tabular}}
\caption{Testing fixed-length summaries ($k \in [1,5]$) for BertExt (first block) and as GSum guidance (second block). Metrics are ROUGE-1/2/L and BERTScore precision~(B\textsubscript{P}) and recall (B\textsubscript{R})}
\label{tab:appendix-fixed-k-testing}
\end{table}

\subsection{Evaluating Guidance Length Prediction}
\label{sec:appendix-oracle-approx}
To predict the length of OracleExt in the variable-length guidance setting, we employ a logistic regression classifier and a BERT-based classifier (cf. \cref{sec:experimental-setup}).
Detailed evaluation results for both classification models are given in \cref{tab:appendix-oracle-approx}.

\begin{table}[t]
\small
\centering
\subfloat[Dataset: MIMIC-CXR]{
\resizebox{\columnwidth}{!}{
\begin{tabular}{lllr}
\toprule
{} & \textbf{LR-Approx} & \textbf{BERT-Approx} & \\
\cmidrule(lr){2-2}
\cmidrule(lr){3-3}
\textbf{Target} & F-1 (Prec./Rec.) & F-1 (Prec./Rec.) & Support \\
\midrule
$k = 0$ & 46.2 (80.0/32.4) & 53.7 (60.0/48.6) & 37 \\
$k = 1$ & 71.1 (63.4/80.9) & 71.7 (68.9/74.9) & 824 \\
$k = 2$ & 39.7 (43.1/36.9) & 46.7 (45.1/48.4) & 512 \\
$k = 3$ & 30.9 (53.3/21.8) & 43.2 (61.5/33.3) & 225 \\
Macro Avg. & 47.0 (59.9/43.0) & 53.9 (58.9/51.3) & 1,598 \\
\midrule
On training set & 52.3 (64.1/47.6) & 62.5 (69.7/58.5) & 122,500 \\
\bottomrule
\end{tabular}}}
\quad
\subfloat[Dataset: OpenI]{
\resizebox{\columnwidth}{!}{
\begin{tabular}{lllr}
\toprule
{} & \textbf{LR-Approx} & \textbf{BERT-Approx} & \\
\cmidrule(lr){2-2}
\cmidrule(lr){3-3}
\textbf{Target} & F-1 (Prec./Rec.) & F-1 (Prec./Rec.) & Support \\
\midrule
$k = 0$ & 77.7 (85.9/70.9) & 84.0 (86.6/81.6) & 103 \\
$k = 1$ & 84.6 (77.2/93.6) & 85.4 (79.8/92.0) & 450 \\
$k = 2$ & 19.8 (36.1/13.7) & 28.4 (39.6/22.1) & 95 \\
$k = 3$ & 15.4 (50.0/9.1) & 8.7 (100.0/4.5) & 22 \\
Macro Avg. & 49.4 (62.3/46.8) & 51.6 (76.5/50.1) & 670 \\
\midrule
On training set & 58.5 (83.3/54.3) & 51.1 (53.0/51.0) & 2,342 \\
\bottomrule
\end{tabular}}}
\caption{
Precision, recall and F1 for length prediction of OracleExt. Scores are provided per class on the test set, and as macro-average for both the training and test set.
Support indicates the number of samples in each class.
}
\label{tab:appendix-oracle-approx}
\end{table}

\subsection{Including the Background Section}
\label{sec:appendix-background-experiment}
\begin{figure}[t]
\includegraphics[width=\columnwidth]{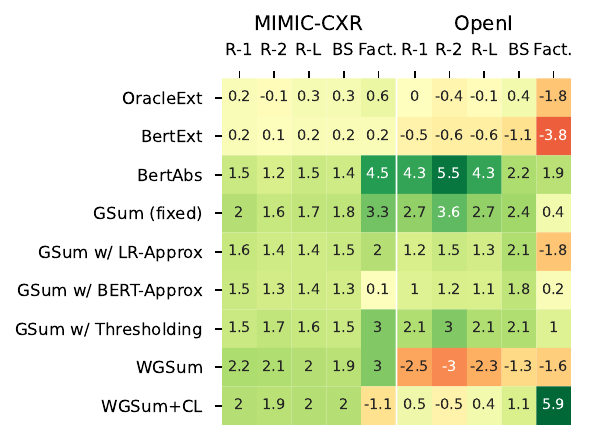}
\caption{Training models to summarize both background and findings improves most abstractive methods. Scores as absolute delta to the same models without the background section (cf. \cref{tab:results}).}
\label{fig:background-experiment}
\end{figure}

To understand to what extent the background section carries important information for summarizing findings to impression, we prepend it to the findings section and retrain all models.
It can be observed that this change improves most abstractive methods on both datasets (\cref{fig:background-experiment}).
For extractive methods results stay largely on par or get worse, indicating that these models do not effectively integrate the background information.

\subsection{Examples of Duplicated Findings and Impressions}
\label{sec:appendix-duplication-examples}
We anecdotally observed a large degree of duplication within MIMIC-CXR which may cause corpus-level inconsistencies (see discussion in~\cref{sec:error-analysis-discussion}).
This section further quantifies the degree of duplication and provides several examples.
Throughout, we only consider instances of \emph{exact} duplication.
Of the 122,500 training reports in MIMIC-CXR, we find that 11.9\% have a findings section occurring more than once.
We present examples of duplicate findings with \emph{different} impressions in~\cref{tab:duplicates}.
In addition, we calculate a \emph{label entropy} over the probabilities that each impression occurs for a given finding.
We posit that duplicate finding-impression pairs may negatively impact model training in two ways.
First, for findings with a high label entropy, the training loss cannot not stabilize (i.e., it is not clear which impression the model should favor).
Second, for findings with a low label entropy, the model may learn a kind of ``majority vote,'' which in turn may render models not sensitive enough to generate useful summaries for slightly different findings.
We leave further investigation of report duplication to future work.

\subsection{Factuality of Additions}
\label{sec:appendix-radnli}
As discussed in~\cref{sec:error-analysis-discussion}, we use RadNLI~\cite{Miura:2021:NAACL} to get a first estimate for the factuality of additions marked by annotators in the error analysis.
RadNLI obtained an accuracy of 77.8\% on a test set of 480 manually labeled sentence pairs in MIMIC-CXR~\cite{Miura:2021:NAACL}, which we consider sufficient for an initial exploration of the factuality of additions.
\cref{tab:radnli-results} presents a breakdown of the RadNLI predictions for all addition spans and models.
It can be seen that the majority of additions is either neutral to the findings section, or entailed by it.
Yet, between 23.4\% and 29.3\% of additions contradict at least one findings sentence, indicating that factuality of radiology report summarization methods can also further be improved.

\subsection{Error Analysis: Responses to \emph{Other} Category}
We analyze the annotators' comments from the \emph{other} error category, and categorize these errors into two-level hierarchy using a bottom-up approach. Our categorization alongside definitions, examples and counts is shown in \cref{tab:app:other}.

\section{Replication Details for Modeling}
\label{sec:appendix-replication-models}
We report hyperparameters of the summarization models in \cref{tab:hyperparameters-summarization}, and for models that predict the length of OracleExt (\textsc{lr-approx}/\textsc{bert-approx}) in \cref{tab:hyperparameters-oracle-approx}.
All models were trained on NVIDIA RTX A6000 GPUs with 48GB of memory.

\begin{table}[t]
\small
\centering
\resizebox{\columnwidth}{!}{
\begin{tabular}{lrrr}
\toprule
\textbf{Model} & \textbf{Entail} & \textbf{Neutral} & \textbf{Contradict}\\
\midrule
BertAbs & 31.9\% & 44.7\% & 23.4\% \\
GSum w/ Thresholding & 34.5\% & 36.2\% & 29.3\% \\
WGSum & 32.0\% & 44.0\% & 24.0\% \\
WGSum+CL & 33.3\% & 41.2\% & 25.5\% \\
\bottomrule
\end{tabular}}
\caption{Factuality of additions in candidates (i.e., spans categorized as ``2a Finding/interpretation''), as per RadNLI~\cite{Miura:2021:NAACL}.}
\label{tab:radnli-results}
\end{table}

\section{Replication Details for Error Analysis}
\label{sec:appendix-replication-error-analysis}

\paragraph{Sample statistics.}
For inclusion in the error analysis, samples were drawn uniformly at random from the official test set of MIMIC-CXR. We compare statistics of the sample with those of the full test set in~\cref{tab:error-analysis-sample-statistics}.

\begin{table}[t]
\small
\centering
\begin{tabular}{lccc}
\toprule
\textbf{Aspect} & \textbf{Full Test Set} & \textbf{Sample} \\
\midrule
Reports & 1,598 & 100 \\
Avg. $|\bm{x}|_{t}$ & 70 {\color{gray} $\pm$ 27.4} & 63 {\color{gray} $\pm$ 20.4} \\
Avg. $|\bm{x}|_{s}$ & 6.2 {\color{gray} $\pm$ 1.9} & 5.7 {\color{gray} $\pm$ 1.6} \\
Avg. $|\bm{y}|_{t}$ & 19 {\color{gray} $\pm$ 15.2} & 18 {\color{gray} $\pm$ 12.4} \\
Avg. $|\bm{y}|_{s}$ & 1.8 {\color{gray} $\pm$ 1.0} & 1.7 {\color{gray} $\pm$ 0.9} \\
Novelty & 69.8\% & 69.7\% \\
CMP & 71.9\% & 70.3\% \\
\bottomrule
\end{tabular}
\caption{Statistics of the MIMIC-CXR test set and the sample used in the error analysis.}
\label{tab:error-analysis-sample-statistics}
\end{table}

\paragraph{Aggregating span-based annotations.}
\label{sec:appendix-span-based-aggregation}
From the three annotations we form a ``gold standard'' as follows: for binary questions we take a majority vote.
For span-based annotations, we first group (partially) overlapping spans, and then take a majority vote within each group.
We provide an example for the majority voting of span-based annotations below. A1, A2, A3, denote annotators, and \texttt{[--eX--]} denotes an error of category X.

\begin{small}
\begin{verbatim}
Tokens:  a   b  c   d   e   f   g   h
A1    : [-e1-]  [-----e2----]
A2    : [-e1-]  [-e1-] [-e2-]
A3    : [-e1-]                 [--e1--]
---------------------------------------
Group :    1          2           3
---------------------------------------
Vote  : [-e1-]         [-e2-]
\end{verbatim}
\end{small}

\paragraph{Inter-annotator agreement (IAA).}
\label{sec:appendix-iaa}
We calculate $F_1$ for span-annotations (\citet{Deleger:2012:AMIA}, categories 1 and 2), and Krippendorffs' Alpha~\cite{Krippendorff:1970:ALPHA} for binary judgments (categories 3 and 4) and report the IAA by category in~\cref{tab:iaa}.
\begin{table*}[t]
\centering
\small
\resizebox{\textwidth}{!}{
\begin{tabular}{llllll}
\toprule
\textbf{Parameter} & \textbf{BertExt} & \textbf{BertAbs} & \textbf{GSum} & \textbf{WGSum} & \textbf{WGSum+CL} \\
\midrule
Training Steps (MIMIC) & 20,000 & 20,000 & 20,000 & 50,000 & 100,000 \\
Training Steps (OpenI) & 20,000 & 20,000 & 20,000 & 20,000 & 20,000 \\
LR (Encoder) & $2\mathrm{e}{-3}$ & $2\mathrm{e}{-4}$ & $2\mathrm{e}{-4}$ & $5\mathrm{e}{-2}$ & $2\mathrm{e}{-4}$ \\
LR (Decoder) & n/a & $2\mathrm{e}{-2}$ & $2\mathrm{e}{-2}$ & $5\mathrm{e}{-2}$ & $5\mathrm{e}{-2}$ \\
Warmup (Encoder) & 10,000 & 20,000 & 20,000 & 8000 & 10,000 \\
Warmup (Decoder) & n/a & 10,000 & 10,000 & 8000 & 7000 \\
Dropout & 0.1 & 0.2 & 0.2 & 0.1 & 0.2 \\
Checkpoint freq. (MIMIC) & 1000 & 2000 & 2000 & 2000 & 2000 \\
Checkpoint freq. (OpenI) & 1000 & 2000 & 2000 & 200 & 200 \\
Decoding & n/a & Beam search & Beam search & Beam search & Beam search \\
Prediction length & n/a & $\ge$ 5 tokens & $\ge$ 5 tokens & $\ge$ 5 tokens & $\ge$ 5 tokens \\
Training GPUs & 3 & 5 & 5 & 4 & 3 \\
Inference GPUs & 1 & 1 & 1 & 1 & 1 \\
Base model & bert-base-uncased & bert-base-uncased & bert-base-uncased & None & \begin{tabular}[x]{@{}l@{}}dmis-lab/biobert-\\base-cased-v1.1\end{tabular} \\
Parameters & 120,512,513 & 180,222,522 & 205,433,914 & 82,260,794 & 221,600,069 \\
\bottomrule
\end{tabular}
}
\caption{Hyperparameters of BertExt/BertAbs~\cite{Liu:2019:EMNLP}, GSum~\cite{Dou:2021:NAACL}, WGSum~\cite{Hu:2021:ACL} and WGSum+CL~\cite{Hu:2022:ACL}. Training steps, warmup and learning rates were adapted as described in \cref{sec:experimental-setup}. Remaining parameters kept as in the original publications.}
\label{tab:hyperparameters-summarization}
\end{table*}

\begin{table}[t]
\centering
\small
\begin{tabular}{lp{0.62\columnwidth}}
\toprule
\textbf{Parameter} & \textbf{Setting} \\\midrule
\emph{LR-Approx} & \\ \midrule
Features & Bag-of-words, unigrams with minimum document-frequency of 5, tf-idf \\
Parameters & 3718 (MIMIC-CXR), 592 (OpenI) \\
Regularization & L2 regularization with strength $C=1$ \\
Solver & SAGA \\
Max. Iterations & 1000 \\\midrule
\emph{BERT-Approx} & \\ \midrule
Checkpoint & \texttt{distilbert-base-cased} \\
Parameters & 65,784,580 \\
Optimizer & Adam \\
Learning rate & $2\mathrm{e}{-5}$ \\
Epochs & 3 \\
Dropout & 0.2 \\
Batch size & 16 \\
Checkpoint freq. & 250 \\
Hardware & 6 GPUs \\ \bottomrule
\end{tabular}
\caption{Hyperparameters for guidance length prediction models.}
\label{tab:hyperparameters-oracle-approx}
\end{table}

\begin{table}[t]
\small\centering
\begin{tabular}{c@{\hspace{1\tabcolsep}}lrr}
\toprule
\textbf{\#} & \textbf{Category} & \textbf{IAA} & \textbf{Count} \\
\midrule
\multicolumn{4}{l}{\emph{Omissions from reference}} \\
1a & Finding/interpretation & 0.64 & 774 \\
1b & Comparison & 0.34 & 236 \\
1c & Ref. to prior report & 0.23 & 43 \\
1d & Communication/followup & 0.83 & 216 \\
\midrule
\multicolumn{2}{l}{Total} & 0.61 & 1269 \\
\addlinespace\multicolumn{4}{l}{\emph{Additions to candidate}} \\
2a & Finding/interpretation  & 0.66 & 718 \\
2b & Comparison  & 0.44 & 155 \\
2c & Ref. to prior report  & 0.08 & 17 \\
2d & Communication/followup  & 0.65 & 72 \\
2e & Contradicting finding & 0.26 & 34 \\
\midrule
\multicolumn{2}{l}{Total} & 0.60 & 996 \\
\addlinespace
3 & Incorrect location & 0.26 & 111 \\
4 & Incorrect severity & 0.41 & 121 \\
\bottomrule
\end{tabular}
\caption{Inter-annotator agreement (IAA) by category and total number of annotations before majority voting.}
\label{tab:iaa}
\end{table}

\begin{table*}[t]
\small\centering
\setlength\tabcolsep{3pt}
\resizebox{\textwidth}{!}{
\begin{tabular}{lp{0.35\textwidth}rcccrp{0.35\textwidth}}
\toprule
\textbf{\#} & \textbf{Finding} & \textbf{Dups.} & \textbf{\%} & $|\bm{y}^*|$ & \textbf{H} & \textbf{Count} & \textbf{Top-5 Impressions} \\
\midrule
\multicolumn{8}{l}{\emph{Most frequent duplicates}} \\\midrule
1 & \multirow[t]{5}{=}{PA and lateral views of the chest provided. There is no focal consolidation, effusion, or pneumothorax. The cardiomediastinal silhouette is normal. Imaged osseous structures are intact. No free air below the right hemidiaphragm is seen.} & 1141 & 0.93 & 26 & 0.12 & 1061 & No acute intrathoracic process.\\
& & & & & & 45 & No acute intrathoracic process\\
& & & & & & 3 & No acute intrathoracic process. \_, MD\\
& & & & & & 3 & No acute intrathoracic process. Specifically, no pneumothorax.\\
& & & & & & 3 & No evidence of pneumonia.\\
\addlinespace
2 & \multirow[t]{5}{=}{Heart size is normal. The mediastinal and hilar contours are normal. The pulmonary vasculature is normal. Lungs are clear. No pleural effusion or pneumothorax is seen. There are no acute osseous abnormalities.} & 1033 & 0.84 & 34 & 0.11 & 974 & No acute cardiopulmonary abnormality.\\
& & & & & & 24 & No evidence of pneumonia.\\
& & & & & & 3 & No radiographic evidence of pneumonia.\\
& & & & & & 2 & No acute cardiopulmonary abnormality. No displaced fracture identified. If there is continued concern for a rib fracture, consider a dedicated rib series.\\
& & & & & & 1 & Improving bibasilar atelectasis and decreasing bilateral effusions.\\
\addlinespace
3 & \multirow[t]{5}{=}{The lungs are clear without focal consolidation. No pleural effusion or pneumothorax is seen. The cardiac and mediastinal silhouettes are unremarkable.} & 753 & 0.61 & 47 & 0.20 & 665 & No acute cardiopulmonary process.\\
& & & & & & 15 & No acute cardiopulmonary process. No focal consolidation to suggest pneumonia.\\
& & & & & & 8 & No pneumonia.\\
& & & & & & 7 & No evidence of pneumonia. No acute cardiopulmonary process.\\
& & & & & & 4 & No acute cardiopulmonary process. No significant interval change.\\
\midrule\multicolumn{8}{l}{\emph{Duplicates with highest impression entropy}} \\\midrule
4 & \multirow[t]{5}{=}{The heart is normal in size. The mediastinal and hilar contours appear within normal limits. There is no pleural effusion or pneumothorax. The lungs appear clear. Bony structures appear within normal limits.} & 25 & 0.02 & 2 & 0.99 & 14 & No evidence of acute cardiopulmonary disease.\\
& & & & & & 11 & No evidence of acute disease.\\
& & & & & & & \\
& & & & & & & \\
& & & & & & & \\
\addlinespace
5 & \multirow[t]{5}{=}{The lungs are clear. There is no pneumothorax. The heart and mediastinum are within normal limits. Regional bones and soft tissues are unremarkable.} & 25 & 0.02 & 2 & 0.94 & 16 & Clear lungs with no evidence of pneumonia.\\
& & & & & & 9 & Clear lungs.\\
& & & & & & & \\
& & & & & & & \\
& & & & & & & \\
\addlinespace
6 & \multirow[t]{5}{=}{The lungs are well expanded and clear. Hila and cardiomediastinal contours and pleural surfaces are normal.} & 23 & 0.02 & 15 & 0.92 & 6 & Normal. No evidence of pneumonia.\\
& & & & & & 2 & No evidence of pneumonia.\\
& & & & & & 2 & Normal chest radiograph.\\
& & & & & & 2 & No pneumonia.\\
& & & & & & 1 & Normal. No evidence of mass.\\
\bottomrule
\end{tabular}}
\caption{Examples of exact duplicates in the training set of MIMIC-CXR. In total, there are 14,596 reports with duplicated findings (11.9\% of the training data). The table shows the number of reports with a given finding (\textbf{Dups.}), the relative frequency in the training set (\textbf{\%}), the number of distinct impressions with this finding ($|\bm{y}^*|$), the entropy over the impression frequencies (\textbf{H}), and the top-5 impressions with their respective \textbf{Count}.}
\label{tab:duplicates}
\end{table*}

\begin{table*}[t]
\small
\centering
\resizebox{\linewidth}{!}{
\begin{tabular}{l p{12em} p{12em} p{12em}  r}
\toprule
\textbf{(Sub-)Category} & \textbf{Description} & \textbf{Example} & \textbf{Explanation} & \textbf{Count} \\
\midrule

\multicolumn{4}{l}{\emph{1. Incorrect findings: the finding in the reference is replaced with a different and incorrect finding.}} & 29 \\ \midrule

Finding &
incorrectness affects the main finding. &
no acute \textcolor{Orange}{intrathoracic} process. &
The reference uses ``\textcolor{Orange}{cardiopulmonary} process'' instead of ``\textcolor{Orange}{intrathoracic} process''. &
21 \\

Past state &
incorrectness affects a past state of the patient. &
\textcolor{Orange}{increased} opacity in the right lung.. &
The reference mentions that the opacity is \textcolor{Orange}{new} and did not exist before.  &
7 \\

Other &
incorrectness affects other aspects.  &
bilateral pleural \textcolor{Orange}{effusions},..., slightly improved... &
The \textcolor{Orange}{improvement} is used to describe a second finding in the reference.   &
1 \\\midrule

\multicolumn{4}{l}{\emph{2. Imprecise findings: the description of the finding or some of its aspects is imprecise compared to the reference.}} & 73 \\ \midrule

Finding &
the description of the finding itself is imprecise compared to the reference.  &
...no \textcolor{Orange}{displaced} fractures are seen. &
The reference uses ``\textcolor{Orange}{acute} fractures'' instead of ``\textcolor{Orange}{displaced} fractures'' (the reference is more general). &
21 \\

Location &
the location of the finding is imprecise.  &
retrocardiac opacity compatible with pneumonia... &
The references specifies the exact location: ``\textcolor{Orange}{Left lower lobe pneumonia}''. &
21 \\

Certainty &
the summary is presented with a different degree of certainty.  &
bilateral middle lobe opacities \textcolor{Orange}{could represent} atelectasis or pneumonia. &
The reference is certain about the finding.  &
9 \\

Repetition &
some findings are repeated. &
unchanged \textcolor{Orange}{bibasilar bronchiectasis} and \textcolor{Orange}{bibasilar bronchiectasis}. &
\textcolor{Orange}{bibasilar bronchiectasis} is mentioned twice. &
6 \\

Count &
the count in the finding is imprise. &
right pleural \textcolor{Orange}{effusion}. &
The reference adds ``Multiloculated'', i.e.,  ``\textcolor{Orange}{Multiloculated} right pleural effusion'' &
2 \\

Size &
the size of the finding is added/omitted/different. &
multiple bilateral pulmonary nodules \textcolor{Orange}{measuring up to 2. 5 cm}. &
The reference omits the size. &
1 \\

Other &
other aspects about the finding are imprecise. &
interval resolution of large right pleural effusion... &
The reference includes other clinical information. &
13 \\\midrule

\multicolumn{4}{l}{\emph{3. Minor/secondary: errors that do not affect the finding.}} & 21 \\\midrule

Limitation &
some limitations of the examination are (not) mentioned.  &
no definite acute cardiopulmonary process. &
The reference adds \textcolor{Orange}{``based on this limited, portable examination''}. &
15 \\

Phone calls &
The time of a telephone call is different.   &
...these findings were discussed with dr. \_ by \_ via \textcolor{Orange}{telephone on \_ at 4 : 45 pm}. &
The reference mentions a different time for the phone call. &
4 \\

Recommendation &
errors related to recommendations.  &
short radiographic follow up is recommended \textcolor{Orange}{within \_ weeks} to document resolution. &
The reference omits \textcolor{Orange}{``within \_ weeks''}.  &
2 \\
\bottomrule
\end{tabular}}
\caption{Bottom-up categorization of errors from the \textit{Other} category with descriptions, examples and counts.}
\label{tab:app:other}
\end{table*}

\clearpage
\clearpage
\section{Annotation Guidelines}
\label{sec:appendix-annotation-guidelines}

\paragraph{Introduction.}
We consider automatic impression generation for English radiology reports of chest imaging examinations. These reports conventionally have three sections (example in \cref{fig:app:report}).

\begin{enumerate}[noitemsep,parsep=0pt,partopsep=0pt]
    \item \textbf{Background.} A description of the exam, patient information, and relevant prior exams.
    \item \textbf{Findings.} A description or itemization of the radiologists' observations based on the radiographs.
    \item \textbf{Impression.} A concise summary of the most important findings, including inferences and any recommendations.
\end{enumerate}

\begin{figure}[h]
\setlength\fboxsep{7pt}
\fbox{%
\parbox{0.95\linewidth}{%
\begin{sanseriffont}
\begin{center} FINAL REPORT \end{center}

EXAMINATION: CHEST (PA AND LAT).\\%

INDICATION: \_\_\_ year old woman with ?pleural effusion  // ?pleural effusion\\%

TECHNIQUE: Chest PA and lateral\\%

COMPARISON: \_\_\_ \\

FINDINGS:\\
Cardiac size cannot be evaluated.  Large left pleural effusion is new.  Small
right effusion is new.  The upper lungs are clear.  Right lower lobe opacities
are better seen in prior CT.  There is no pneumothorax.  There are mild
degenerative changes in the thoracic spine \\

IMPRESSION: Large left pleural effusion
\end{sanseriffont}
}}
\caption{Example radiology report.}
\label{fig:app:report}
\end{figure}

\paragraph{Study setup.}
We are looking to determine typical failures that automatic impression generation systems make. You will be shown a reference impression and four candidate impressions. The reference was written by a radiologist, and the candidates were generated by automatic systems. Your goal is to judge the accuracy of candidate impressions \emph{based on a reference impression}. For each candidate, you will be asked to identify any errors that it may have.

\paragraph{Definition of ``error.''} We define an error as a difference between the candidate and reference. An error can be one of the following:

\begin{enumerate}[noitemsep,parsep=0pt,partopsep=0pt]
\item Omissions
    \begin{enumerate}[(a),noitemsep,topsep=0pt,parsep=0pt,partopsep=0pt]
        \item Omission of finding/interpretation
        \item Omission of comparison describing a change from a previous examination
        \item Omission of reference to prior report while making a comparison
        \item Omission of next steps, recommendation, communications
    \end{enumerate}
\item Additions
    \begin{enumerate}[(a),noitemsep,topsep=0pt,parsep=0pt,partopsep=0pt]
        \item Additional finding/interpretation
        \item Mention a comparison that is not present in reference
        \item Additional reference to prior report while making a comparison
        \item Additional mention of next steps, recommendation, communications
        \item Additional finding/interpretation which contradicts reference
    \end{enumerate}
\item Incorrect location/position of finding
\item Incorrect severity of finding
\item Other difference between candidate and reference (please describe...)
\end{enumerate}
\cref{tab:app:guidelines:examples} shows an example for each error category.

\begin{figure*}[t]
\centering
\includegraphics[width=\textwidth]{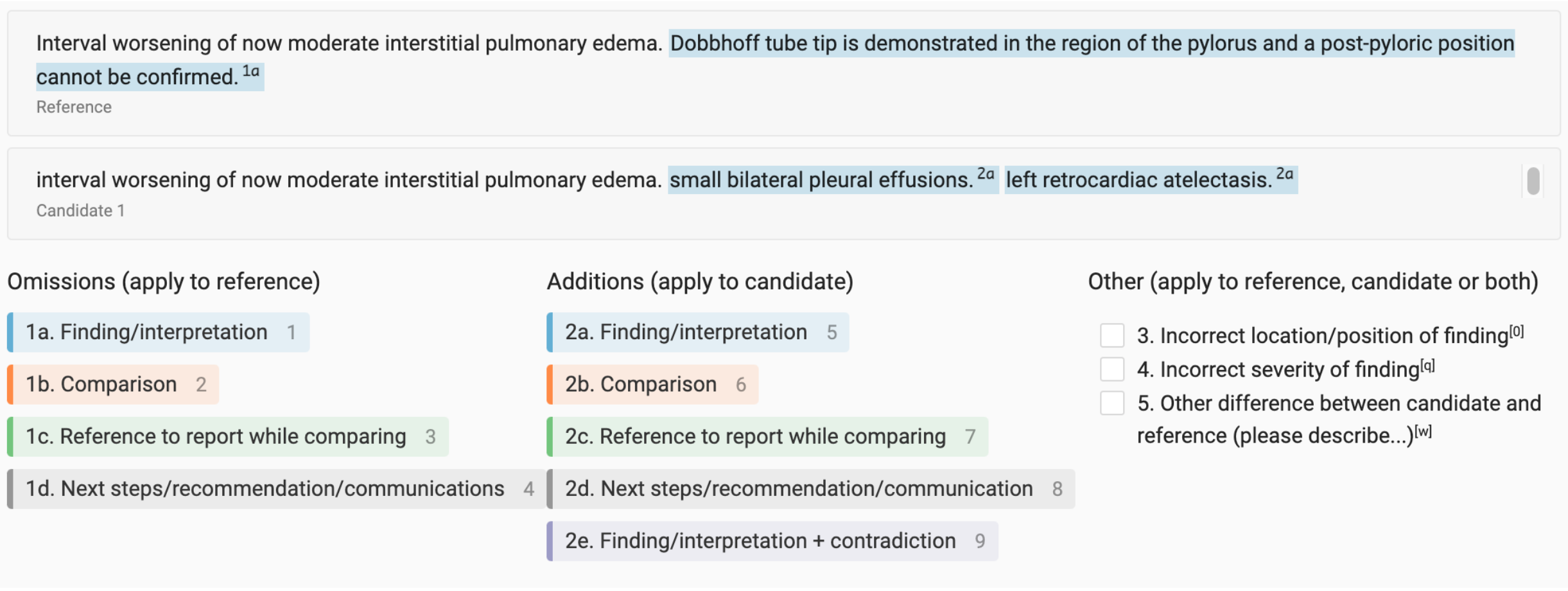}
\caption{A candidate with two additional findings. Even though they are placed next to each other in the text, apply the category \textcolor{purple}{2a. Additional finding/interpretation} twice.}
\label{fig:app:guidelines:ex1}
\end{figure*}

\paragraph{Annotating errors as inline annotations.}
You are asked to annotate errors as \emph{inline annotations}. For each error that you identify, first select the error category and then highlight the relevant text snippet with your mouse. This applies the category. If you have to remove an annotation, press on the highlight and use your backspace/delete key (\keystroke{Entf} or \keystroke{\textleftarrow Backspace}). If one of the above categories occurs multiple times, please annotate all of them \emph{separately} (see \cref{fig:app:guidelines:ex1}). Some general guidelines:

\begin{itemize}[noitemsep]
\item A candidate may have multiple errors, so please add all that apply.
\item Some candidates will be the same, so please assign the same errors to all candidates.
\item For additional findings that are plausible, pick \textcolor{purple}{2a. Additional finding/interpretation}. In the context of the full report, these additions may be correct. What this category aims to capture is that the system included information which the radiologist chose not to include. If a finding contradicts the reference, select \textcolor{purple}{2e. Additional finding/interpretation which contradicts reference}.
\item Use \textcolor{purple}{5. Other} freely, especially if you find it difficult to assign any of the above categories. These remarks help us to better understand and characterize potential errors.
\item You can ignore differences in word choice if they are synonymous. Example: \emph{``may reflect developing consolidation''} is equal to \emph{``could represent early consolidation.''}
\end{itemize}
Finally, always use your best judgment when assessing the reports. If you are in doubt, you can add any questions/comments about the report or the error categories in the given box.

\subsection*{Corner Cases}
\paragraph{How to annotate ``3. Incorrect location/position of finding'' and ``4. Incorrect severity of finding''?}
Only apply if both reference and candidate mention a finding, \emph{and} when there is a mismatch in severity/location. In the example below, both mention effusion, but the reference does not specify the size of effusion, whereas the candidate states that there are ``small'' effusions. Therefore, apply \textcolor{purple}{4. Incorrect severity of finding}.

\noindent\fbox{%
\parbox{0.98\linewidth}{%
\begin{sanseriffont}
\textbf{Reference:}
interval worsening of now moderate interstitial pulmonary edema. bilateral pleural effusions.\\
\textbf{Candidate:}
interval worsening of now moderate interstitial pulmonary edema. small bilateral pleural effusions.
\end{sanseriffont}
}}

\paragraph{Opacities vs. consolidation.}
Often, opacities are used in place of consolidation and vice versa. In those cases, apply \textcolor{purple}{5. Other} with a comment similar to \textcolor{purple}{``opacities not equal consolidation, but otherwise correct''}.

\noindent\fbox{%
\parbox{0.98\linewidth}{%
\begin{sanseriffont}
\textbf{Reference:}
Improved right lower medial lung peribronchial consolidation. \\
\textbf{Candidate:}
right lower medial lung peribronchial opacities have improved.
\end{sanseriffont}
}}

\paragraph{No acute abnormality vs. COPD.}
Does ``no acute abnormality'' contradict ``COPD''? No, for the purposes of our evaluation, COPD is not an \emph{acute} disease, so this is not contradicting. In the example below, following categories apply: (1) ``COPD'' is missing $\rightarrow$ \textcolor{purple}{1a. Omission of finding/interpretation}, (2) ``opacity is resolved'' $\rightarrow$ \textcolor{purple}{1b. Omission of comparison describing a change from a previous examination}, (3) ``no acute cardiopulmonary abnormality'' $\rightarrow$  \textcolor{purple}{2a. Additional finding/interpretation}.

\noindent\fbox{%
\parbox{0.98\linewidth}{%
\begin{sanseriffont}
\textbf{Reference:}
Left basilar opacity is resolved. COPD. \\
\textbf{Candidate:}
no acute cardiopulmonary abnormality.
\end{sanseriffont}
}}

\paragraph{Misleading grammar or sentence structure.}
In general, disregard grammatical errors. However, please pay attention to any \emph{logical flaws} that arise because of grammar errors or a misleading sentence structure. In the example below, the \emph{``and''} in the candidate implies that both ``bronchiectasis'' and ``peribronchial consolidation'' have improved, whereas the reference only states that the consolidation has improved. In those cases, apply \textcolor{purple}{5. Other} and add a comment similar to \textcolor{purple}{``logical error because of grammar.''}

\noindent\fbox{%
\parbox{0.95\linewidth}{%
\begin{sanseriffont}
\textbf{Reference:}
Bilateral lower lung bronchiectasis with improved peribronchial consolidation\\
\textbf{Candidate:}
bilateral lower lung bronchiectasis and peribronchial consolidation have improved since \_.
\end{sanseriffont}
}}

\begin{table*}[t]
\small
\centering
\resizebox{\textwidth}{!}{
\begin{tabular}{p{10em} p{13em} p{13em} p{12em} }
\toprule
\textbf{Error} & \textbf{Reference} & \textbf{Candidate} & \textbf{Explanation} \\
\midrule

\multicolumn{4}{l}{\emph{Omissions (apply to reference)}}\\\midrule
1a. Omission of finding/interpretation &
New left lower lobe infiltrate \hlomma{and effusion}. &
New left lower lobe infiltrate.	&
Effusion is missing. \\\addlinespace

1b. Omission of comparison describing a change from a previous examination &
\hlommb{In comparison to \_ exam, there is interval near-complete resolution of bilateral pleural effusion.}	&
No evidence of acute cardiopulmonary process. &
Resolution of effusion is not described, therefore the comparison is missing. \\\addlinespace

1c. Omission of reference to prior report while making a comparison &
Increased pulmonary edema \hlommc{compared to \_.}	&
increased pulmonary edema.	&
While the candidate correctly states that the edema has increased, it lacks the reference to the prior report (or the date of it). \\\addlinespace

1d. Omission of next steps / recommendation / communications &
No pneumothorax or pneumomediastinum. \hlommd{Recommend repeat PA and lateral imaging later today to verify these findings.} Otherwise unremarkable chest radiograph. \hlommd{These findings were communicated to Dr. \_ at 11:55 a.m. by telephone by Dr. \_.}	&
No pneumothorax or pneumomediastinum. &
The candidate does not include the followup \emph{(recommend repeat PA)} and the remark about a communication with another doctor \emph{(These findings were communicated [...])}. \\\addlinespace

\midrule
\multicolumn{4}{l}{\emph{Additions (apply to candidate)}}\\
\midrule
2a. Additional finding / interpretation	&
Slight increased hazy opacities at the right lung base which may reflect developing consolidation.	&
slightly increased hazy opacities at the right lung base which may represent \hlomma{atelectasis or} developing consolidation. &
Atelectasis is not mentioned in the reference. This finding is not contradicting the reference. It may be correct in the context of the full report. \emph{Same as 1a, but in the other direction.} \\\addlinespace

2b. Mention a comparison that is not present in reference	&
Mild to moderate pulmonary edema, increased from \_.	&
Mild to moderate pulmonary edema, increased from \_. \hlommb{Stable cardiomegaly.} &
``Stable'' suggests that the state of a finding was compared to a previous examination. This comparison is not made in the reference. \emph{Same as 1b, but in the other direction.} \\\addlinespace

2c. Additional reference to prior report while making a comparison	&
&
&
\emph{Same as 1c, but in the other direction.}\\\addlinespace

2d. Additional mention of next steps / recommendation / communications	&
&
&
\emph{Same as 1d, but in the other direction.}\\\addlinespace

2e. Additional finding / interpretation which contradicts reference &
Unchanged size and position of right-sided hydropneumothorax. &
\hladde{Development of new} right-sided hydropneumothorax &
Unchanged vs. development of new \\\addlinespace

\midrule
\multicolumn{4}{l}{\emph{Incorrect location, Incorrect Severity, Other}}\\
\midrule
3. Incorrect location/position of finding &
New \hlommb{left} lower lobe infiltrate	&
New \hlommb{right} lower lobe infiltrate	&
Left vs. right \\\addlinespace

4. Incorrect severity of finding &
In comparison prior exam, \hlommb{there is near-complete resolution} of bilateral pleural effusion &
In comparison to \_ exam, \hlommb{there is resolution} of bilateral pleural effusion	&
Near complete vs. resolved \\\addlinespace

5. Other &
Slight increased hazy \hlommb{opacities} at the right lung base which may reflect developing consolidation &
Slight increased hazy \hlommb{opacity} at the right lung base which may reflect developing consolidation &
Difference in multiplicity
\\\addlinespace

5. Other &
left picc terminates within the \hlommb{upper} svc.	&
left picc terminates within the \hlommb{proximal} svc. &
Ambiguous location \\\addlinespace

5. Other &
No acute abnormalities identified to explain patient's cough \hlommb{and asthma flare}. &
no acute abnormalities identified to explain patient's cough. &
Asthma flare is a symptom, which was not mentioned in the candidate. \\
\bottomrule
\end{tabular}
}
\caption{Examples for all error categories.}
\label{tab:app:guidelines:examples}
\end{table*}

\end{document}